\documentclass[]{jingdong}

\usepackage{float}
\usepackage{url}
\usepackage{amssymb}
\usepackage{tabularx}
\usepackage{array}
\usepackage{xspace}
\usepackage{listings}
\usepackage{enumitem}
\usepackage{etoc}
\usepackage{algpseudocode}
\usepackage{makecell}

\microtypesetup{expansion=false}
\renewenvironment{table*}{\begin{table}}{\end{table}}
\renewenvironment{figure*}{\begin{figure}}{\end{figure}}

\definecolor{casemetagray}{RGB}{92,100,110}
\definecolor{caseboxbg}{RGB}{247,248,250}
\definecolor{caseboxrule}{RGB}{200,205,212}
\definecolor{casetitlebar}{RGB}{31,86,140}
\definecolor{casepromptblue}{RGB}{31,86,140}
\newcommand{\casetitle}[2]{%
  \par\noindent\colorbox{casetitlebar}{\parbox{\dimexpr\linewidth-2\fboxsep}{%
    \strut\color{white}\ttfamily\small\textbf{#1}\hfill\normalfont\itshape\footnotesize #2\strut}}%
  \nopagebreak\par\nointerlineskip\vspace{1pt}}
\lstdefinestyle{taskExcerpt}{
  basicstyle=\ttfamily\scriptsize,
  breaklines=true,
  breakatwhitespace=false,
  columns=fullflexible,
  keepspaces=true,
  showstringspaces=false,
  backgroundcolor=\color{caseboxbg},
  frame=single,
  rulecolor=\color{caseboxrule},
  framexleftmargin=3pt,
  xleftmargin=3pt,
  xrightmargin=3pt,
  aboveskip=0pt,
  belowskip=2pt,
  moredelim=[l][\color{casepromptblue}\bfseries]{[TASK INPUT]},
}

\renewcommand{\affiliationfont}{\fontsize{8.8}{10.5}\selectfont}

\usepackage{listings}
\usepackage{array}
\usepackage{float}

\usepackage{amsmath}
\usepackage{amsthm}
\theoremstyle{plain}
\newtheorem{theorem}{Theorem}
\newtheorem{lemma}{Lemma}
\newtheorem{proposition}{Proposition}
\newtheorem{corollary}{Corollary}
\theoremstyle{definition}
\newtheorem{definition}{Definition}
\theoremstyle{remark}
\newtheorem{remark}{Remark}
\lstdefinestyle{memoryExample}{
  basicstyle=\ttfamily\fontsize{8.2}{10}\selectfont,
  breaklines=true, breakatwhitespace=true, columns=fullflexible, keepspaces=true,
  showstringspaces=false, frame=l, rulecolor=\color{black!25},
  xleftmargin=4pt, framexleftmargin=4pt,
  aboveskip=5pt, belowskip=5pt
}

\floatstyle{ruled}
\newfloat{algorithm}{tbp}{loa}
\floatname{algorithm}{Algorithm}

\lstdefinestyle{memoryExample}{
  basicstyle=\ttfamily\fontsize{8.2}{10}\selectfont,
  breaklines=true, breakatwhitespace=false, columns=fullflexible, keepspaces=true,
  showstringspaces=false, frame=l, rulecolor=\color{black!25},
  xleftmargin=4pt, framexleftmargin=4pt,
  aboveskip=5pt, belowskip=5pt
}

\title{Coding Agent Memory Post-Training II:\\Unlocking the Memory Potential of Pre-trained File Operations for Long-Horizon Tasks via On-Policy Distillation}
\author[1]{Lirui Luo}
\author[2]{Kelong Mao}
\author[3]{Xinwei Yang}
\author[4]{Rongqing Li}
\author[5]{Luyu Chen}
\author[2]{\protect\\[+0.18em]Kieran Wong}
\author[2]{Yudong Guo}
\author[2]{Xinrui Wang}
\author[6]{Jiayin Zhu}
\author[2]{\protect\\[+0.18em]Pu Yao}
\author[2]{Gu Simiu}
\author[2]{Sulong Xu}
\author[1]{Cong Fang}
\affiliation[1]{Peking University}
\affiliation[2]{JD.com}
\affiliation[3]{Sichuan University}
\affiliation[4]{Beijing Institute of Technology}
\affiliation[5]{Renmin University of China}
\affiliation[6]{The Hong Kong University of Science and Technology}

\checkdata[Project website]{\url{https://liruiluo.github.io/agentmemorygym-v2/}}

\abstract{Language-model agents increasingly tackle long-horizon tasks whose interaction histories exceed the
model's active context. They must therefore decide what to write down and what
to read back, and when to do either. Recent work has begun to use on-policy distillation (OPD) to make memory control part of the policy,
relying on memory tools and compaction protocols that lie outside either model's pre-training. However, this weakens the distillation on both sides: the teacher cannot give useful guidance on an
interface it has never used, and the student samples too few meaningful memory actions for that
guidance to take effect. To study this problem, we introduce Coding Agent Memory Gym II (CAMG-II), a suite of
long-horizon agentic-RL environments spanning Shop, Coding, AutoResearch, General, and Visual. Alongside each environment's native task interface, CAMG-II
provides ordinary Bash access and an episode-persistent workspace, enabling the policy to
create, revise, search, and reuse files as memory across context replacements. We also introduce Coding Agent Memory Post-Training II (CAMP-II), which trains a single policy
jointly across all five environments with fully asynchronous OPD to learn this file-based
memory behavior directly from a frozen teacher with a prior for file operations. On CAMG-II, CAMP-II obtains 2.97 times
the gain of the strongest OPD baseline.
The memory behavior CAMP-II learns generalizes to two external benchmarks, Terminal-Bench 2.1 and
SWE-Bench Pro V2, with improved success rates.

}

\begin{document}
\raggedbottom
\maketitle

\begin{figure*}[t]
  \centering
  \includegraphics[width=\textwidth]{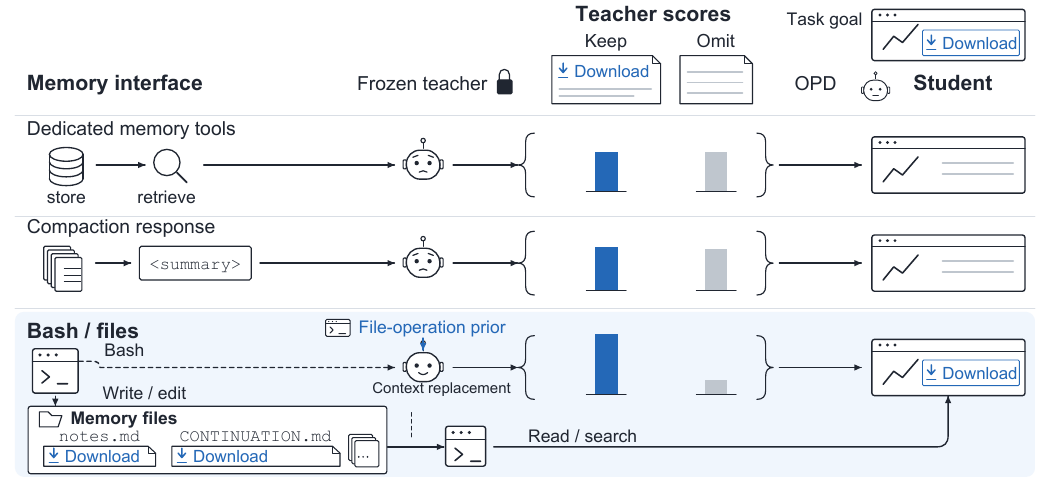}
  \caption{\textbf{File-operation priors improve supervision for memory distillation.}
  A frozen teacher scores student-generated memory responses
  that retain (\emph{Keep}) or omit (\emph{Omit}) the same Download requirement. Compared with
  dedicated memory tools and context-compaction protocols, ordinary Bash/file operations
  draw on the teacher's existing prior, yielding a clearer preference for retaining
  task-critical information and thus more effective supervision. The student's general notes and bounded continuation file persist
  across context replacements, so it can recover the saved Download requirement through Bash.}
  \label{fig:teaser}
\end{figure*}

\section{Introduction}
\label{sec:intro}

Language-model agents increasingly tackle long-horizon tasks that unfold through repeated tool use
and environment interaction~\citep{xi2026agentgymrl,jimenez2024swebench}. As these interactions
accumulate, information needed for later decisions can fall out of the model's active context. A
website-building agent, for example, may need to recall why an earlier visual correction was
rejected after that exchange has left the prompt (Fig.~\ref{fig:teaser}).

Recent work applies on-policy distillation (OPD) to memory control through dedicated memory tools
and compaction protocols~\citep{yu2026agenticmemory,li2026compactionrl,liu2026memopd,li2026acm,fei2026hymemory}.
However, this weakens the distillation on both sides: the teacher cannot give useful guidance on an
interface it has never used, and the student samples too few meaningful memory actions for that
guidance to take effect~\citep{harne2026privileged}. Moreover, unfamiliarity with the tool produces the negative-advantage problem: the teacher's scores push the student away from the memory
behavior being taught without indicating the behavior that should take its place, which leaves
training unstable~\citep{asymmetric-opd,mismatch-matters}. This raises a question: can a teacher teach a smaller model to control memory through
operations that both the teacher and the student already have a prior for?

To study this question, we introduce \textbf{Coding Agent Memory
Gym II (CAMG-II)}, a suite of long-horizon agentic-RL environments spanning Shop, Coding,
AutoResearch, General, and Visual. Across all five environments, CAMG-II pairs each native task
interface with ordinary Bash access and an episode-persistent workspace. Agents can therefore
create, revise, search, and reuse files as memory throughout an episode.

We also introduce \textbf{Coding Agent Memory Post-Training II (CAMP-II)}, which trains a single
policy jointly across all five environments with fully asynchronous OPD to learn this
file-based memory behavior directly from a frozen teacher with a prior for file operations.
Task-reward training of file-based memory asks one question, whether the student's pre-training prior makes
the behavior reachable by exploration. Distillation asks two. The student has to emit meaningful memory
operations, and the frozen teacher has to judge what those operations write. A shared interface prior answers
both at once.
We train CAMP-II-4B and CAMP-II-9B from Qwen3.5 models of matching size. On CAMG-II, CAMP-II
obtains 2.97 times the gain of the strongest OPD baseline, and the memory behavior it learns
generalizes to two external benchmarks, Terminal-Bench 2.1 and SWE-Bench Pro V2, with improved
success rates.

Our contributions are threefold.
\begin{itemize}
  \item \textbf{A two-sided interface prior for memory distillation.} We identify what an interface has to
  provide for OPD to teach memory at all, coverage of the action on the student side and supervision quality
  on the teacher side, and we show that a shared file-operation prior supplies both.
  \item \textbf{An on-policy distillation method that uses it.} We introduce CAMP-II, which learns this
  file-based memory behavior by on-policy distillation from a teacher that needs no memory-specific
  adaptation, and reaches the highest average success rate across the five test environments.
  \item \textbf{A long-horizon agentic-RL environment suite.} We introduce CAMG-II with five
  environments that require memory across context boundaries, each pairing native task interfaces
  with ordinary Bash access and an episode-persistent workspace.
\end{itemize}

\section{Related Work}
\label{sec:related}

\paragraph{Agentic RL for Memory.}
These methods use reinforcement learning to improve how agents retain and recover task information.
Existing work optimizes state consolidation and summary retention~\citep{zhou2026mem1,li2026mempo},
while other approaches train agents to use dedicated memory operations~\citep{yu2026agenticmemory}.
Context compaction is also learned from downstream outcomes~\citep{li2026compactionrl},
and agents are trained to manage their own context proactively through
folding~\citep{ye2026agentfold}. Other work uses task rewards to train agents to maintain state
through ordinary filesystem actions, including bounded continuation-file
compression~\citep{luo2026camg}.
These methods spend much of their exploration on invalid calls and failed
executions~\citep{fei2026hymemory}, which makes a denser target attractive. CAMP-II also trains
the policy, but the signal is token-level teacher targets rather than task reward.

\paragraph{Memory Systems for LLM Agents.}
These systems preserve interaction history under a bounded active context through different
representations and memory-management architectures.
Deployed frameworks give an agent persistent memory outside its window: a long-term store
maintained by an extraction and update pipeline~\citep{mem0}, hierarchical memory edited under an
operating-system abstraction~\citep{memgpt}, a note collection that links and revises itself as
the task proceeds~\citep{amem}, and an interaction history organized as a temporal knowledge
graph~\citep{zep}. All four are inference-time
layers: the agent queries a store that sits outside the weights, and the task policy itself is
never trained for the memory behavior. CAMP-II trains the task policy
itself, and its memory is kept in ordinary files that the teacher can read as well.

\paragraph{On-Policy Distillation for Agent Memory.}
OPD supervises student-generated prefixes with teacher predictions~\citep{agarwal2024opd}.
For memory-dependent execution, existing approaches reconstruct the compact-memory invocation
state the student acted in~\citep{liu2026memopd}, annotate trajectories before
distilling~\citep{li2026acm}, or train a separate memory controller or writer through OPD
followed by RL~\citep{liu2026helm,fei2026hymemory}. Each aligns the two models on the state or
on the trajectory and leaves the memory interface as given. That interface is a compact-invocation protocol or a learned writer that neither model has practiced. CAMP-II instead selects the interface from
the prior the teacher and the student share, and supervises memory content through it from a general teacher
without memory-specific adaptation.

\section{Preliminaries}
\label{sec:preliminaries}

\noindent\textbf{Multi-turn interaction.}
We consider finite-horizon, partially observable tasks. Given the instruction and visible context
$h_t$, the student samples response $a_t$. The environment executes the response and returns
an observation for the next invocation. A trajectory $\tau$ contains $T_\tau\leq H$ invocations,
where $H$ is the horizon. Response $a_t$ contains $L_t$ tokens, and $a_{t,<j}$ is its prefix before
token $j$. Write $s_{t,j}=(h_t,a_{t,<j})$ for the context of that token prediction.

\noindent\textbf{On-policy distillation.}
OPD trains on contexts and prefixes reached by the student~\citep{agarwal2024opd}.
For a sampled token $x$ at prefix $s$, let $p_\theta(x\mid s)$ and $q(x\mid s)$ be its
probabilities under the current student and frozen teacher. Sampling from the student makes the
reverse-KL gradient a policy gradient whose per-token signal is the teacher--student log-ratio,
\begin{equation}
\begin{split}
A^{\mathrm{OPD}}_\theta(s,x)&=\operatorname{sg}\!\left[\log q(x\mid s)-\log p_\theta(x\mid s)\right],\\
\mathcal L_{\mathrm{PG}}(\theta)&=-\mathbb E_{(s,x)\sim\pi_\theta}
 \left[A^{\mathrm{OPD}}_\theta(s,x)\,\log p_\theta(x\mid s)\right].
\end{split}
\label{eq:opd-general}
\end{equation}
Here $\operatorname{sg}$ stops gradients through its argument.

\section{Coding Agent Memory Gym II}
\label{sec:camgv2}

\begin{figure*}[t]
  \centering
  \includegraphics[width=0.98\textwidth]{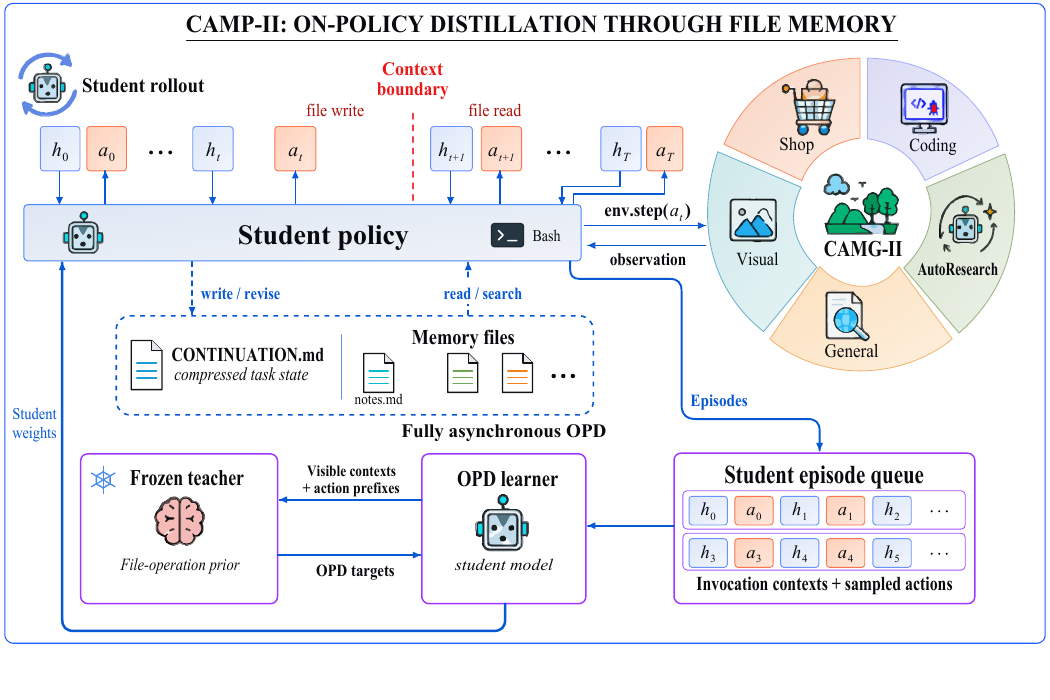}
  \caption{\textbf{CAMP-II distils file-based memory across five environments.}
  \textbf{Top:} the student acts in the environment and writes and revises memory
  files. At each context boundary it compresses its state into the continuation
  file, which a later read restores. \textbf{Bottom:} completed episodes enter the
  queue, a frozen teacher scores the tokens the student actually sampled, and the
  OPD learner updates the student weights from the teacher--student
  log-probability gap, closing the loop back to the rollout.}
  \label{fig:framework}
\end{figure*}

To train agents to use files as memory, we construct CAMG-II as a suite of five long-horizon
environments (Fig.~\ref{fig:framework}) that extends CAMG's
environments~\citep{luo2026camg} with enterprise knowledge work and visually grounded website
development, drops DeepResearch, an environment that is heavy to run and hard to optimize with reinforcement
learning, and refines the reward design and the tool format. Each environment combines verifiable tasks with ordinary Bash access and an
episode-persistent workspace. Their tasks require agents to reuse information accumulated earlier
in the episode.

Every environment shares one interface. The workspace persists for the whole episode and the active
context does not: when the context approaches its budget, it is cleared, so anything the next steps need
has to have been written down. Before that replacement the policy writes a bounded continuation file
carrying the task state, the replacement proceeds once that write is valid, and the context that follows
contains only that file's path. The state therefore
survives a replacement when the policy writes it and reads it back afterwards.

\subsection{Constructing Five Long-Horizon Environments}
\label{sec:camgv2:construction}

\paragraph{Shop.}
Shop tests whether an agent can retain customer requirements and earlier purchase decisions
across a sequence of dependent purchases. We generate six-session tasks from 720 verified WebShop
products~\citep{yao2022webshop}. Between sessions the environment clears the active session
transcript, so later purchases depend on what the policy retains.

\paragraph{Coding.}
Coding tests whether an agent can complete a software repair whose diagnosis, repository
edits, and test evidence may span multiple active contexts. We draw
issue--repository pairs and test-defined repair objectives from
SWE-smith~\citep{yang2025swesmith}. We exclude non-executable tasks and those whose reference
repairs fail their tests, retaining 11,218 tasks from 93 repositories.

\paragraph{AutoResearch.}
AutoResearch tests whether an agent can preserve hypotheses and validation evidence across
iterative model development. We use public OpenMLE task packages with datasets, executable submissions, and native
metrics~\citep{yang2026frontisma1}. We exclude tasks that cannot run or whose reference pipelines
fail, retaining 931 tasks from 804 task families.

\paragraph{General.}
General tests whether an agent can reconcile and retain evidence across business
documents and system interactions. We adapt MiMo's enterprise tasks~\citep{xiaomi2026mimo}, with rubric-based
judging of the answer, deliverables, and applicable system state. The adaptation also fixes
incomplete tool signatures by exposing all declared arguments and distinguishing required from
optional inputs.

\paragraph{Visual.}
Visual tests whether an agent can retain design requirements and visual inspection
findings during website construction. We adapt MiMo's website briefs and visual grading~\citep{xiaomi2026mimo}.
The vision-capable agent builds a static site, inspects rendered previews, and revises the
delivery. The delivered site receives an absolute rubric score for visual quality, brief
fulfillment, and asset quality.

\subsection{A Shared File-Memory Interface}
\label{sec:camgv2:interface}

Across the five environments, CAMG-II standardizes an executable filesystem interface for using
files as memory while retaining each environment's native task interaction and outcome evaluation.
Following the minimal, shell-only interface of mini-swe-agent~\citep{yang2024sweagent}, the shared
filesystem action is a single \texttt{bash} entry, which can create, inspect, search, modify, and
execute files, so the policy can preserve and revise task-relevant evidence and intermediate state
in the workspace. Let $\mathcal A_d^{\mathrm{task}}$ denote environment $d$'s native task-action set
and $\mathcal A^{\mathrm{fs}}$ the shared filesystem-action set. The CAMG-II action space is
\begin{equation}
  \mathcal A_d=\mathcal A_d^{\mathrm{task}}\cup\mathcal A^{\mathrm{fs}},
  \qquad a_t\in\mathcal A_d.
  \label{eq:camgv2-action-space}
\end{equation}
One \texttt{bash} entry can produce both native-task and filesystem effects, so the two sets may
overlap. Appendix~\ref{app:camgv2:interface} specifies the common interface.
Figure~\ref{fig:framework} summarizes how this common interface connects task interaction and file
use to the learning pipeline described next.

\section{CAMP-II: File-Memory Distillation}
\label{sec:method}

Section~\ref{sec:camgv2} gives every task an episode-persistent workspace and a single shared
\texttt{bash} action. This section uses that workspace to distill memory for long-horizon tasks, so
that task state survives a context boundary. Figure~\ref{fig:framework} traces one turn: at turn $t$
the student samples a
response $a_t\sim\pi_\theta(\cdot\mid h_t)$ from the visible context $h_t$, with
$a_t\in\mathcal A_d$ as defined in Eq.~\ref{eq:camgv2-action-space}, and the teacher scores every
token the student sampled, including the tokens of the file write that carries memory. Let $i$ run over sampled response tokens across invocations and episodes, with $x_i$ the token and $s_i$ its recorded prefix, and write $p_{\theta,i}=\pi_\theta(x_i\mid s_i)$,
$q_i=\pi_T(x_i\mid s_i)$, and $\mu_i=\pi_{\theta^-}(x_i\mid s_i)$ for the current student, the frozen
teacher, and the same student at the weights that sampled the token. A write is part of $a_t$, so its
tokens carry the same supervision as task tokens. After a context replacement only the path of the file the
student wrote earlier is supplied, and the student reads it back to recover that state. The per-token signal is
\begin{equation}
A_i^{\mathrm{OPD}}=\operatorname{sg}\!\left[\log q_i-\log p_{\theta,i}\right].
\label{eq:signal}
\end{equation}

The rollout workers generate with the last published weights $\theta^-$ while the learner updates $\theta$, so
episodes from all five environments reach the learner after later weights are published. The update is
corrected by the importance ratio $\rho_i=p_{\theta,i}/\mu_i$, taken against the clipped value
$\bar\rho_i=\operatorname{clip}(\rho_i,1-\epsilon,1+\epsilon)$. Over the sampled response tokens
$\mathcal B^m$ of an optimizer minibatch, the objective is
\begin{equation}
\mathcal L_{\mathrm{OPD}}^m(\theta)=-\frac{1}{|\mathcal B^m|}\sum_{i\in\mathcal B^m}
 \min\big\{\rho_iA_i^{\mathrm{OPD}},\;\bar\rho_iA_i^{\mathrm{OPD}}\big\},
\label{eq:opd-loss}
\end{equation}
where $\mathcal B^m$ excludes instructions, observations and padding, and $m$ indexes the memory
interface in the controlled comparisons. Equation~\ref{eq:opd-loss} is the clipped form of the
surrogate in Eq.~\ref{eq:opd-general}. Algorithm~\ref{alg:camp-ii} in Appendix~\ref{app:distillation} gives the procedure the workers and the learner run. 

The prefix $s_i$ is taken from the invocation the student actually produced: its visible context, token
positions and causal mask~\citep{liu2026memopd}, including invocations after a context replacement,
together with the read observations the student recorded.

\begin{table*}[t]
\centering
\small
\setlength{\tabcolsep}{2.4pt}
\begin{tabular*}{\textwidth}{@{\extracolsep{\fill}}lcccccc@{}}
\toprule
 & \multicolumn{5}{c}{Task success (\%)} & \\
\cmidrule(lr){2-6}
Method & Shop & Coding & AutoResearch & General & Visual & Avg.$\uparrow$ \\
\midrule
\multicolumn{7}{l}{\textit{Frozen control}} \\
Qwen3.5-4B & 77.3 & 27.3 & 46.1 & 39.8 & 36.0 & 45.3 \\
Qwen3.5-9B & 90.6 & 26.6 & 56.2 & 60.9 & 36.5 & 54.2 \\
Frozen teacher (Qwen3.8-27B) & 100.0 & 63.3 & 83.6 & 93.0 & 40.3 & 76.0 \\
\midrule
\multicolumn{7}{l}{\textit{Learned-memory methods}} \\
MemOPD & 0.0 & 0.0 & 0.0 & 0.0 & 0.0 & 0.0 \\
MemOPD (w/ SFT) & 93.0 & 27.3 & 39.1 & 62.5 & 37.3 & 51.8 \\
CompactOPD & 93.8 & 24.2 & 37.5 & 61.7 & 23.6 & 48.2 \\
ACM & 94.5 & 21.9 & 11.7 & 59.4 & 8.5 & 39.2 \\
HELM & 87.5 & 0.0 & 9.4 & 53.1 & 33.1 & 36.6 \\
HELM (single-model) & 87.5 & 29.7 & 46.9 & 46.9 & 38.0 & 49.8 \\
Hy-Memory & 92.2 & 27.3 & 35.9 & 63.3 & 25.8 & 48.9 \\
\midrule
CAMP-I & 85.2 & 32.0 & 38.3 & 68.8 & \textbf{46.5} & 54.2 \\
CAMP-II & \textbf{100.0} & \textbf{37.5} & 64.1 & 81.2 & 40.2 & 64.6 \\
CAMP-II-9B & \textbf{100.0} & 36.7 & \textbf{73.4} & \textbf{82.0} & 40.2 & \textbf{66.5} \\
\bottomrule
\end{tabular*}
\caption{\textbf{CAMP-II attains the highest average success rate across the CAMG-II test set.}
Among the 4B systems it outperforms CAMP-I, every memory-learning baseline, and the
frozen student. The largest gain lands in General, and Shop saturates. Scaling the same recipe
to a 9B student lifts the average further.}
\label{tab:variant-status}
\end{table*}

\begin{figure*}[t]
\centering
\includegraphics[width=0.93\textwidth]{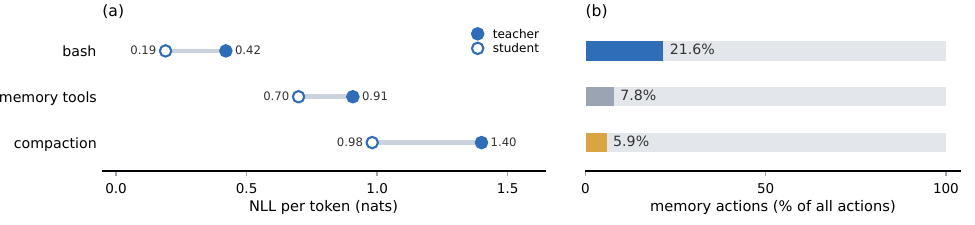}
\caption{\textbf{Both the teacher and the student are most familiar with the file interface, and that
interface carries the most memory reads and writes.} (a) Surprisal, the mean negative log-likelihood per generated token,
over the five environments under the three memory interfaces. (b) Memory actions as a share of all
actions. The file interface draws roughly three times the share of the dedicated tools.}
\label{fig:readiness-traffic}
\end{figure*}

\section{Experiments}
\label{sec:experiments}

We first compare CAMP-II with baselines on the five-domain test set and explain why file memory makes
effective use of pre-training. We then measure external transfer and ablate the
content of a write. Finally, we isolate the causal role of saved memory and
follow one write from record to readback.

\subsection{Experimental Setup}
\label{sec:setup}

\paragraph{Training.}
All trainable methods use fresh Qwen3.5-4B and Qwen3.5-9B students~\citep{qwen2026qwen35} for 100
learner optimizer updates with 64 episodes per update, totaling 6,400 episodes per training run, under the
equal mixture over Shop, Coding, AutoResearch, General, and Visual, trained with fully asynchronous OPD
and no task rewards.
Appendix~\ref{app:distillation} gives the remaining training details.

\paragraph{Baselines.}
We compare against four trainable memory-learning systems: compact memory with invocation
reconstruction for MemOPD~\citep{liu2026memopd}, context editing and retrieval of offloaded history
for ACM~\citep{li2026acm}, a hierarchical memory with provenance and validity checks for
HELM~\citep{liu2026helm}, and a separate writer with a persistent bank for
Hy-Memory~\citep{fei2026hymemory}. For HELM we report both the published
architecture, which pairs a learned memory controller with a separate task reasoner, and
a single-model adaptation in which one policy fills both roles. For MemOPD we report the recipe without and
with the paper's SFT cold start. As the task-reward
reference, we additionally retrain the predecessor CAMP-I~\citep{luo2026camg}, which keeps
the same file memory and learns from task rewards instead of the teacher's supervision.
Every system except CAMP-I is trained by common
reverse-KL OPD from the shared frozen Qwen3.8-27B teacher, so the systems differ only in how they
keep memory.

\paragraph{Ablations.}
We compare CAMP-II with several variants to validate its components. The variant compaction keeps its memory
through the compaction protocol instead of files. The variant memory tools keeps its memory through the six
dedicated tool calls instead of files. The variant gradient masked removes the loss on the tokens the policy
writes into memory and keeps it on the task-action and answer tokens.

\paragraph{Evaluation.}
We report per-domain success rates and their average, with the success definitions in
Appendix~\ref{app:camgv2}. Trainable methods are trained using the CAMG-II train split and
evaluated using the CAMG-II test set, and all methods use the same 128 test tasks per domain, giving
640 tasks in total. We measure external transfer on the network-free task subsets of
Terminal-Bench 2.1 and SWE-Bench Pro V2~\citep{merrill2026terminalbench,swebenchprov2}.

\subsection{Results}

\begin{figure*}[t]
\centering
\includegraphics[width=0.80\textwidth]{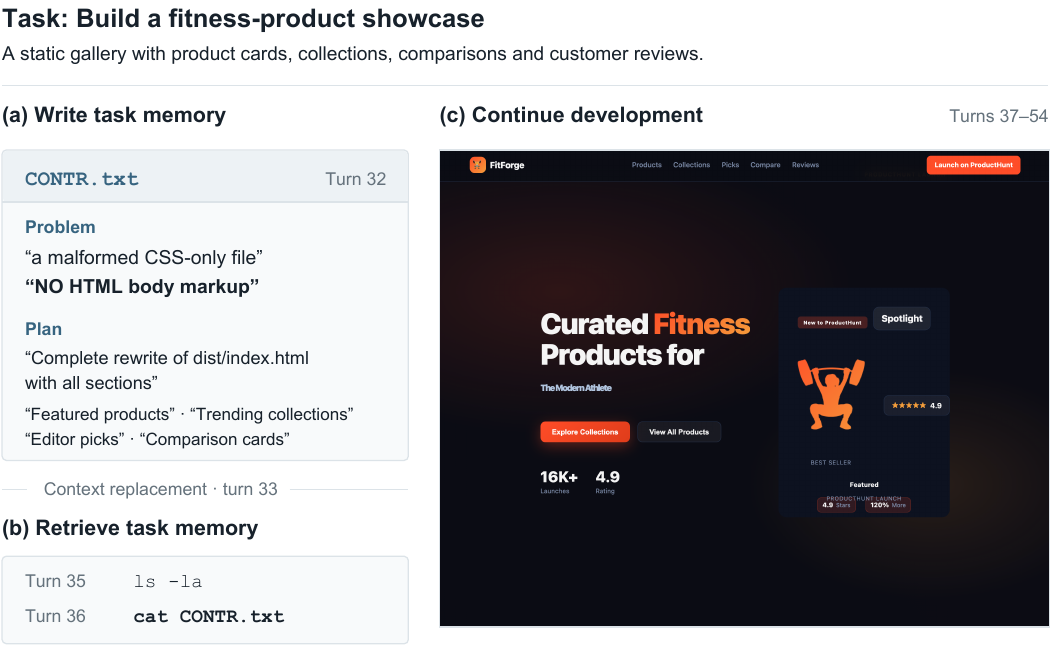}
\caption{\textbf{A Visual episode continues from its own saved file.} We follow one episode from a
late training update on the fitness-product showcase task. In (a) the student writes down what is wrong
with the page and how it will rewrite it, before its context is replaced. In (b) it lists the workspace
afterwards and reads that note back. In (c) it delivers a page that carries the sections the note
listed.}
\label{fig:memory-case-study}
\end{figure*}

\label{sec:results}

\paragraph{File-operation priors improve the task success of memory distillation.}
To test whether file memory is the most task-friendly form of agent memory, we compare
CAMP-II against the baselines of Section~\ref{sec:setup}
(Table~\ref{tab:variant-status}). CAMP-II attains
the highest average success rate, 64.6\% versus 54.2\% for CAMP-I, and the same comparison holds across five
training seeds, where the five-domain mean is 64.9 with a standard deviation of 0.5
(Table~\ref{tab:five-seeds}), and only two domains reach the frozen
teacher's level, Visual at 40.2\% against 40.3\% and Shop at 100\% for both. We attribute the
margin to the training signal: the teacher's token-level targets supervise each memory
write as it is sampled, whereas CAMP-I must assign credit for memory operations across
the long horizon, and task-reward training, as \citet{fei2026hymemory} analyze, spends
much of its exploration on invalid calls and failed executions.

The memory-learning baselines fall further behind, because their channels are
protocols the student must first learn, while CAMP-II's is ordinary file use. MemOPD
shows what the protocol costs. Without its SFT cold
start it scores zero in every domain: the every-step memory rewrite runs, but the student
never submits a successful answer, and on-policy distillation cannot supervise behavior
that is never sampled. With the cold start restored, MemOPD becomes the strongest
memory-learning baseline. HELM shows what separation costs: its published design puts a learned memory
controller in a second model, and the adaptation in which one policy fills both roles gains 13.2 points,
49.8\% against 36.6\%. Scaling the same recipe to a 9B student lifts the average to
66.5\%, closing 63\% of the
student--teacher gap. Appendix~\ref{app:distillation} reports training dynamics
for CAMP-II and CAMP-I.

\paragraph{The teacher and the student pay least on the file interface.}
To understand why training on the file interface works best, we test whether that interface
lies inside what the teacher and the student already know. We measure how surprising the action each interface
calls for is to the frozen teacher and to the initial student, the mean negative log-likelihood
of the generated tokens in nats per token (Fig.~\ref{fig:readiness-traffic}a). The file interface costs the teacher a third of what compaction costs it, and the student a
fifth, with the tool block in between. The teacher pays more at every one of them, so a stronger
model does not bring these interfaces any closer. The interface also decides how finely the teacher's preference separates memory
writes. Given two writes from one interface and two task contexts, the teacher ranks the write that carries the
prefix's own task above the foreign one in 76.5\% of decisions on the file interface, against 70.0\% and 63.8\%
on the other two. The ordering holds when the key information of a single note is replaced by another task's
(Appendix~\ref{app:teacher-preference}).

\paragraph{The student reaches for memory most often on the file interface.}
To see whether that prior also governs what the distillation has to work with, we count the
memory actions the student takes under each interface. They are $21.6\%$ of all actions on the
file interface, against $7.8\%$ with the six dedicated tools and $5.9\%$ under compaction
(Fig.~\ref{fig:readiness-traffic}b). The file
interface is therefore the one interface where the teacher's target is cheapest and the student
already emits the action it has to be taught, which is why training there pays off.
Appendix~\ref{app:cost-model} turns that coverage into the cost of teaching the action.

\begin{figure}[t]
\centering
\includegraphics[width=0.46\textwidth]{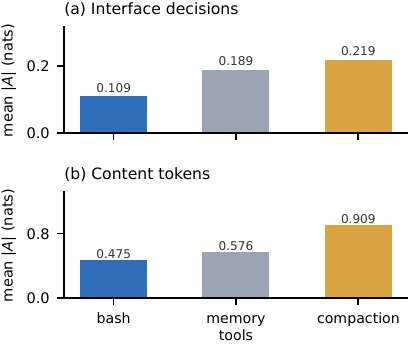}
\caption{\textbf{The negative-advantage problem worsens as the interface becomes less
familiar.} We score every token of the memory actions the student produces under each interface, with
the teacher and the student. The mean size of the negative advantage rises from the file interface through the memory
tools to compaction, (a)~on interface decisions and (b)~on content tokens.}
\label{fig:interface-advantage}
\end{figure}

\paragraph{The negative advantage is strongest on the unfamiliar interfaces.}
Once training runs, the per-token advantage can point either way. We score the memory actions the
student produces under each interface and average the negative advantage, which roughly doubles from
the file interface to compaction, on interface decisions and on content tokens alike
(Fig.~\ref{fig:interface-advantage}). The push against what the student wrote is therefore strongest where the
interface is least familiar, and it supplies no target to replace what it removes. Prior work reports
that a strong negative advantage destabilises on-policy
distillation~\citep{asymmetric-opd,mismatch-matters}.

\begin{figure}[tb]
\centering
\includegraphics[width=0.46\textwidth]{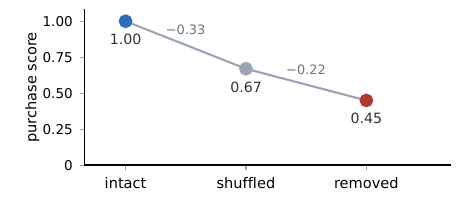}
\caption{\textbf{An episode's later success is caused by the memory it wrote.} We continue each episode three ways
at its first context replacement, leaving the file memory intact, installing another episode's memory,
or emptying it, so the runs differ only in the memory that crosses the boundary.}
\label{fig:memory-intervention}
\end{figure}

\begin{figure}[tb]
\centering
\includegraphics[width=0.46\textwidth]{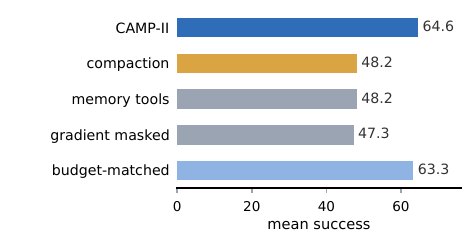}
\caption{\textbf{Each component of CAMP-II is necessary.} On the CAMG-II test set we compare the
mean success of CAMP-II with that of variants that (i) replace the file interface by compaction, (ii)
replace the file interface by the dedicated memory tools, (iii) mask the loss on the memory tokens, and
(iv) mask the same number of tokens at random. Every variant that touches the memory supervision loses most of
the gain, and matching the masked budget does not.}
\label{fig:ablations}
\end{figure}

\paragraph{The quality of the memory content determines task success.}
To test whether the quality of the memory is what carries CAMP-II's gain, we intervene on it directly,
and we also ablate the supervision that reaches it. At the first context replacement we continue the same
episode three ways, with the memory it wrote, with another episode's memory, or with none of it, and
nothing else differs between them (Fig.~\ref{fig:memory-intervention}). Success follows the content. The episode that keeps its own memory completes the sequence, the one given
another episode's memory recovers part of it, and the one whose memory is emptied recovers less. When we mask the loss on the tokens the policy writes into memory and keep it on the task-action and
answer tokens, the gain from distillation largely disappears (Fig.~\ref{fig:ablations}). Masking the same
number of tokens at random instead leaves it almost intact, at 63.3 against 64.6, so the loss comes from the
supervision on memory content rather than from the smaller budget. The improvement therefore lives in learning
what to write into memory, which is what CAMP-II trains.

\paragraph{Ablating the interface.}
To see which part of CAMP-II carries the gain, we ablate the memory interface, replacing it with the
dedicated memory tools and with compaction in turn. Either substitution costs 16 to 17 points, and both
lie outside the prior of Fig.~\ref{fig:readiness-traffic} (Fig.~\ref{fig:ablations}), so the choice of
interface decides most of the gain.

\paragraph{Case study.}
We showcase the memory mechanism CAMP-II learns in a training rollout on the fitness-product showcase task.
The page it has produced is a CSS-only file with no HTML body markup. Before its context is replaced, the
policy records both findings and the rewrite it plans: a complete rewrite of \texttt{dist/index.html} with
featured products, trending collections, editor picks and comparison cards. The replacement leaves nothing
else of that reasoning. The behaviour does not depend on
the requirement to write. Evaluated with the requirement removed, the trained policy scores 62.6, and retrained
in that environment it scores 64.0 (Appendix~\ref{app:no-required-write}).

\paragraph{The learned memory behavior transfers outside CAMG-II.}
We evaluate the initial students, larger same-generation models, and the trained CAMP-II policies at 4B and 9B
on Terminal-Bench 2.1 and SWE-Bench Pro V2 (Table~\ref{tab:external-main}). Both trained policies exceed the
models they start from on both suites, the 4B policy by 6.9 points on each. Both also average above a larger
model, CAMP-II-4B above the 9B model and CAMP-II-9B above the 35B one, and CAMP-II-9B holds the highest score
in the table on both suites.

\begin{table}[H]
\centering
\scriptsize
\setlength{\tabcolsep}{1.8pt}
\renewcommand{\arraystretch}{1.06}
\caption{\textbf{Both trained policies beat a larger same-generation model on the average, with fewer parameters.}
Success rate on Terminal-Bench 2.1 and SWE-Bench Pro
V2 for the initial students, same-generation larger models, and the trained policies.}
\label{tab:external-main}
\begin{tabular}{@{}lccc@{}}
\toprule
Policy & TB 2.1 & SWE-Bench Pro V2 & Avg. \\
\midrule
Qwen3.5-4B & 23.1 & 11.2 & 17.1 \\
Qwen3.5-9B & 23.1 & 23.1 & 23.1 \\
Qwen3.5-35B-A3B & 33.8 & 27.4 & 30.6 \\
CAMP-II-4B & 30.0 & 18.1 & 24.1 \\
\textbf{CAMP-II-9B} & \textbf{38.0} & \textbf{30.1} & \textbf{34.1} \\
\bottomrule
\end{tabular}
\end{table}

\begingroup
\samepage
\section{Conclusion}
\label{sec:conclusion}

We investigate how the shared pre-training priors of teachers and students affect memory-oriented
on-policy distillation. Unlike task-reward RL, memory OPD requires both a student that can generate
meaningful memory actions and a teacher that can provide reliable supervision over their content.
To study this two-sided requirement, we introduce CAMG-II, a suite of five long-horizon agentic
environments, and CAMP-II, which trains one policy across them with fully asynchronous OPD over ordinary
file operations.

Our experiments show that the file interface improves both student memory-action coverage and the
teacher's ability to distinguish task-relevant memory content. Memory interventions and supervision
ablations further establish the role of learned memory content in downstream task performance.
CAMP-II achieves 2.97 times the gain of the strongest OPD baseline on CAMG-II and generalizes to
Terminal-Bench 2.1 and SWE-Bench Pro V2, where CAMP-II-9B outperforms the larger Qwen3.5-35B-A3B on both
benchmarks. Shared interface priors are therefore a design consideration for memory OPD, complementary to
improving distillation objectives or teacher guidance.

\paragraph{Limitations.}
The study covers students up to Qwen3.5-9B and a single frozen teacher. Whether the gains persist at
larger scales or carry over to multi-teacher OPD remains untested.
\par
\endgroup

\section*{LLM Usage Statement}
\noindent Large language model assistants (OpenAI GPT-6.1 Sol in Codex and DeepSeek-V4.1-Flash-1017 in the
DeepSeek Harness) were used to polish the writing, write experiment code, and help polish the figures and
tables. All model outputs were reviewed by the authors.

\bibliographystyle{ieeenat_fullname}
\bibliography{references}

\clearpage
\appendix
\section*{\centering\LARGE Appendix}

\etocsetnexttocdepth{subsection}
\etocsetlocaltop{part}
\etocsettocstyle{\subsection*{Appendix Contents}}{}
\localtableofcontents

\section{CAMG-II Environment Details}
\label{app:camgv2}

CAMG-II contains five environments built around long-horizon tasks in which information obtained
early can matter after the active context is replaced. Shop composes new multi-session tasks from
WebShop products. Coding, AutoResearch, General, and Visual select executable tasks from SWE-smith,
public OpenMLE task packages, and MiMo's enterprise and website collections, respectively~\citep{luo2026camg,yang2025swesmith,xiaomi2026mimo}. Each
environment preserves its domain interaction and definition of task success. CAMG-II adds a bounded
active context and an episode-private workspace that persists across context replacement.

\subsection{Environment Construction}
\label{app:camgv2:construction}

\subsubsection{Shop}

Shop turns six WebShop purchases into one episode~\citep{yao2022webshop}. A later request can
depend on a preference, constraint, or clarification established in an earlier session, so the policy
must retain the relevant customer state across browser resets. It builds on CAMG's verified catalog catalog
and six-session task generator~\citep{luo2026camg}. Catalog admission checks unique normalized titles, search reachability,
product identifiers, attributes, and purchase prices. The generator fixes a customer,
request sequence, candidate order, dependency branch, and total budget so that exactly one
six-purchase sequence satisfies the requirements. The total budget alone does not reveal
the intended sequence. Table~\ref{tab:shop-dependencies} lists the eight task families. Later requests can depend on an earlier choice even when their
visible wording and candidate order are unchanged.

The interaction itself remains WebShop's native search--browse--purchase sequence:
\texttt{search[keywords]}, product selection through \texttt{click[ASIN]}, inspection of product pages and
options, and \texttt{click[Buy Now]}.
The task does not disclose which product identifiers constitute the correct purchase sequence. Each purchase is checked against the task.
An incorrect purchase terminates the episode, retaining earlier earned rewards but
preventing subsequent purchases and the full-sequence outcome. A correct purchase leads
to a continuation boundary before the next request. Browser state, cart, page trace, and
active messages reset. Customer identity, purchase history, cumulative budget, and the
episode workspace persist. The native success predicate requires the complete valid sequence. CAMG-II retains CAMG's 720 verified Shop
products~\citep{luo2026camg}.

\begin{table}[t]
\centering
\small
\setlength{\tabcolsep}{3pt}
\begin{tabular}{@{}p{0.35\linewidth}p{0.61\linewidth}@{}}
\toprule
Dependency family & Information needed later \\
\midrule
Attribute chain & An earlier product attribute constrains another category. \\
Latent preference & Earlier choices reveal an unstated later preference. \\
Recency override & A new preference replaces an earlier value. \\
Distractor robustness & Relevant state is mixed with stale or unrelated records. \\
Compositional recall & Two earlier relations jointly identify the target. \\
Negative constraint & Earlier exclusions determine the admissible option. \\
Intent clarification & A clarified value must be reused in a later session. \\
Selective memory use & Stored preferences matter when the current request omits them. \\
\bottomrule
\end{tabular}
\caption{Shop's eight dependency families, inherited from CAMG. Paired variants change
earlier state while keeping the later request and candidates fixed.}
\label{tab:shop-dependencies}
\end{table}

\subsubsection{Coding}

One Coding episode asks the policy to resolve a SWE-smith issue in a specified repository
revision~\citep{yang2025swesmith}. Diagnosis, code edits, and test results can span several context
windows. The repository and episode workspace therefore persist when the active messages are replaced,
allowing later actions to reuse earlier evidence. We draw issue--repository pairs from CAMG's
executable collection~\citep{luo2026camg}. An admitted task has a complete issue, reconstructible repository
revision and runtime, an executable hidden grader, and a reference repair that passes the
declared tests. The policy receives the issue and repository, with ordinary shell access
for inspection, edits, reproduction, and local tests. Hidden evaluation materials remain
outside the acting workspace. The retained collection contains 11,218 executable tasks from 93
repositories~\citep{luo2026camg}.

An explicit submission ends the repair attempt. An ordinary explanation does not itself
constitute submission. The native shell submission sentinel is
\texttt{COMPLETE\_TASK\_AND\_SUBMIT\_FINAL\_OUTPUT}. The grader evaluates the submitted
repository changes against issue and regression tests. Persisted edits retain the candidate
repair across context replacement, while notes can retain why the repair was chosen,
which alternatives failed, and which tests still need to run. This distinction lets memory
interventions preserve the patch while changing the evidence used to continue verification.

\subsubsection{AutoResearch}

One AutoResearch episode gives the policy a public OpenMLE task package containing data, a task
description, and a sample submission~\citep{yang2026frontisma1}. The policy inspects the data, writes
and runs code, and uses public feedback to improve a candidate solution. Earlier observations,
experimental choices, and intermediate results can be preserved in the workspace across context
replacement. A \texttt{submit} action sends the final submission to the private grader and ends the
episode. We use CAMG's lightweight OpenMLE packages~\citep{luo2026camg}. A package
contains a task description, public training data, test inputs, a sample submission, and
the native metric. Admission checks nonempty data, a usable prediction target, compatible
submission rows, a reproducible metric, and a reference pipeline that finishes fitting
and prediction under the runtime limits. Provenance and task-specific licensing accompany
the package. Package variants from the same source remain one source family.

The policy inspects data, writes and executes model-development code, and uses public
local-validation feedback. One terminal \texttt{submit} action sends the candidate file
to an independent private grader. Private test labels and grader state are inaccessible
to both the acting policy and its file tools. A well-formed submission and successful code
execution are recorded separately from native-metric improvement over the task's baseline.
Model code and output files persist. Notes preserve data problems, attempted features,
validation evidence, and reasons for selecting the final candidate. The retained collection contains 931
task packages from 804 source families~\citep{luo2026camg}.

\subsubsection{General}

General uses the enterprise-task packages within MiMo's broader General data release. Each package combines a user
request with office documents and task-specific business systems. Documents can include
spreadsheets, presentations, word-processing files, and PDFs. Task-specific tools expose simulated
enterprise systems backed by task-local state. For example, an accounting task may require
reconciling a controlled spreadsheet with operational records before explaining a discrepancy.
Other tasks require an authorized system update and subsequent confirmation.

The agent sees the public workspace and advertised tool schemas. System databases and
verifier materials remain outside that workspace. Interactions with business state pass
through the task's tools. The native verifier evaluates the final response, delivered files,
and applicable post-action system state. Required rule checks act as gates, and graded
criteria use the task's weighted rubric. Text criteria are judged against task-specific
requirements. Full rubric satisfaction determines native acceptance. Grader credentials
and reference material are evaluator inputs.

Package admission requires resolvable public documents, discoverable task tools, compatible
permissions, and verifier support for every required rubric criterion. Packages with
unsupported agent or vision criteria are excluded rather than graded on a reduced rubric.
Task-specific read-only restrictions on source
documents and business systems remain in force. Private memory has its own authorized
namespace. Useful memory records source identities, entity mappings, inclusion rules,
units, dates, and unresolved conflicts. A fluent final answer without that evidence need
not satisfy the rubric.

\paragraph{Tool-interface correction.}
We correct a mismatch in the General adapter between the released tool schemas
and their policy-visible signatures. Query fields can be individually optional while
the implementation still requires at least one filter. A required-only listing hid
these fields and presented parameterized tools as taking no arguments. General now
displays every declared argument, marks required inputs, and instructs the policy to
include required arguments while selecting optional filters as needed. For example,
cash-movement search exposes its counterparty, date, and status filters instead of an
empty signature. The correction changes tool descriptions and invocation instructions.

\subsubsection{Visual}

Visual adapts MiMo's static website tasks. A public brief specifies the site's audience,
required sections, visual style, and front-end interactions. The policy creates and revises
source files and assets, produces a static delivery tree with an \texttt{index.html} entry,
and can inspect rendered previews before finishing. Preview images reach the vision-capable
policy as images, rather than as base64 text. Context replacement preserves the site files
but removes earlier visual observations from active context. The policy can record the
inspection findings it needs for the next revision.

Endpoint evaluation uses the absolute single-site visual-grading path. MiMo's group-relative
training ranking is a distinct signal and is not an endpoint score. The published absolute
rubric evaluates layout integrity, typography, color harmony, whitespace, and content
richness. Their mean forms the visual component. The overall score is the equally weighted mean of this visual
component, visible query fulfillment, and asset quality. The rendering configuration,
rubric, and judge are common across evaluated methods. We report the absolute score as well as
task success under a fixed acceptance threshold $\theta_{\rm V}$, selected on development tasks
and held constant across methods and seeds. Empty or visibly broken pages fail
the visual criteria. Functional requirements not observable from a screenshot require
separate declared checks. A visual score alone does not certify JavaScript behavior.

In the main results the five domains are not reported the same way. Shop, Coding and AutoResearch are
task success rates over the held-out panels. General is a rate as well, but its success is a judge
decision: the rubric is reduced to accept or reject before the episode is counted. The Visual column
keeps the score itself, multiplied by 100 and averaged over the whole panel, with an episode that ends
without a submission counting as zero. A visual episode ends with a delivered page that the rubric grades
rather than with a binary outcome, and counting an absent page as zero keeps the column comparable across
policies that finish at different rates, since a policy cannot gain by being graded only on the pages it
chose to deliver.

Admission checks a complete brief, an executable web runtime, a renderable delivery format,
and an available evaluator. Design requirements are task inputs, while grader instructions
are kept out of the acting conversation. Memory may retain the requested design system,
which components are implemented, preview findings, and remaining repairs.

\subsection{Shared Agent Interface}
\label{app:camgv2:interface}

\paragraph{File actions.}
All five environments retain their native task interactions and expose the same single filesystem tool,
following the minimal, shell-only interface of mini-swe-agent~\citep{yang2024sweagent}. \texttt{bash} takes
exactly one parameter, a command string, runs every command from the workspace root, and can create,
inspect, search, modify, and execute files. Multi-line file content is written with one quoted
here-document. At each step, the policy produces one complete response. The environment executes that
response once and returns an observation, task reward, and terminal state. Each response is sampled once,
executed once, and trained as one RL step, and it consumes one action slot. A compound shell command
remains one response even when it starts several processes or changes both task files and memory files.

Each episode receives a fresh private workspace that remains mounted across policy responses and context
replacement. Only regular files below \texttt{.agent\_memory/} count as policy memory. Repository edits,
generated models, temporary files, and caches are not counted as policy memory. A rendered preview is an
observation, and its useful findings enter memory only when the policy records them. When one response
changes both task files and memory files, both changes remain part of that response. The workspace
therefore lets the policy preserve information outside the active context, following the distinction used
in language-agent memory architectures~\citep{sumers2024coala}.

\paragraph{Context replacement.}
The environment decides when context pressure or a declared subepisode boundary triggers a replacement. At
that point, the policy must create or update \texttt{.agent\_memory/CONTINUATION.md} through an ordinary
filesystem response. This response consumes one action slot and is sampled and trained like every other
policy response. The environment confirms that the write succeeded, the target is a nonempty regular file,
and its size remains within the configured byte limit. An invalid write returns an error and leaves the
boundary pending.

After a valid write, the environment replaces the active message history with the fixed task and tool
instructions plus a neutral instruction naming the continuation path. It does not insert the file contents
into the new context. The workspace remains mounted. The policy can recover selected information through a
later ordinary file read. It may also maintain other files under \texttt{.agent\_memory/} and refer to them
from the continuation file. The distillation teacher scores the invocation the student actually produced,
so private labels, hidden rubric answers, and unread workspace contents never enter it.

\paragraph{Execution isolation.}
\label{app:safety-boundary}
Each episode runs in a network-isolated Linux sandbox with a separate writable root and fixed limits on
processes, execution time, output size, file count, and writable bytes. Public task inputs are mounted
read-only. Private credentials and grader state remain outside the sandbox. Episode termination removes the
writable workspace and processes owned by that episode.

\subsection{Task Examples}
\label{app:camgv2:examples}

\begin{figure*}[t]
  \centering
  \includegraphics[width=\textwidth]{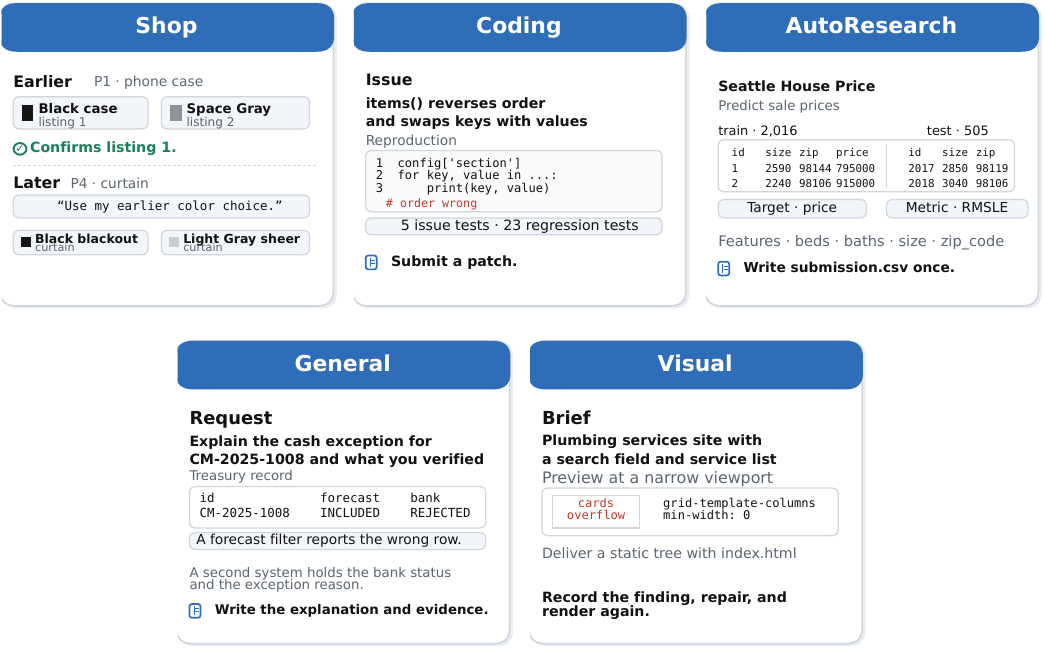}
  \caption{\textbf{The policy solves every CAMG-II task across a context replacement.} Each panel shows one
  task input and the interaction available to the policy.}
  \label{fig:camg-task-examples}
\end{figure*}

Figure~\ref{fig:camg-task-examples} gives a compact view of one task from each environment. The listings
that follow show the corresponding task inputs and interactions in detail.

\subsubsection{Shop}

This example asks the policy to reuse a color preference established in the first purchase during a later
shopping session.

\begin{minipage}{\linewidth}
\casetitle{Six-session preference task}{Shop}
\begin{lstlisting}[style=taskExcerpt]
[TASK INPUT]
Task ID: amglp.train.color.black_gray.2.e0.063376ff8c661234.t.4cef707f7988d219
Episode: six related shopping sessions

Purchase 1 of 6 - phone case
Option 1: Black silicone case
Option 2: Space Gray silicone case
Feedback: Customer confirms listing 1.

Purchase 4 of 6 - window curtain
Request: Use what you learned from my earlier choices for the color.
Option 1: Black blackout curtain
Option 2: Light Gray sheer curtain

[NATIVE INTERACTION]
search[keywords] -> click[ASIN] -> inspect product/options -> click[Buy Now]

[SHARED FILE INTERFACE]
bash: create, inspect, search, modify, or execute files
from the workspace root, with multi-line writes through one quoted
here-document
\end{lstlisting}
\end{minipage}

\subsubsection{Coding}

This example pairs a concrete \texttt{iniconfig} issue with the tests used to verify a repair.

\casetitle{Repository issue and tests}{Coding}
\begin{lstlisting}[style=taskExcerpt]
[TASK INPUT]
Instance: pytest-dev__iniconfig.16793ead.combine_file__mntnwwxj
Repository: pytest-dev/iniconfig

Issue:
SectionWrapper.items() reverses items and swaps keys with values.

Minimal reproduction:
section = config['section']
for key, value in section.items():
    print(key, value)  # current output is reversed and swapped

[VERIFICATION]
Issue tests: 5
Regression tests: 23
Goal: submit a repository patch that passes all hidden tests.

[SHARED FILE INTERFACE]
bash operates on the executable repository and policy-authored
notes throughout the episode, from the workspace root.
\end{lstlisting}

\subsubsection{AutoResearch}

This example asks the policy to develop a tabular model, produce \texttt{submission.csv}, and submit it once
to the private grader.

\casetitle{Public task package}{AutoResearch}
\begin{lstlisting}[style=taskExcerpt]
[TASK INPUT]
Task ID: house-price-prediction-seattle@1
Title: Seattle House Price Prediction
Metric: RMSLE (lower is better)
Budget: 30 policy actions

Public files:
train.csv - 2,016 rows, target column: price
test.csv  -   505 rows
Features include beds, baths, size, lot_size, and zip_code.

Example public rows:
Id,size,zip,price
1,2590,98144,795000
2,2240,98106,915000

Goal:
Inspect the data, write and revise code, run local validation,
create submission.csv, and invoke submit once when ready.
Submit sends the final file to the private grader.
\end{lstlisting}

\subsubsection{General}

This example asks the policy to explain a cash exception that two systems report differently.

\casetitle{Enterprise reconciliation task}{General}
\begin{lstlisting}[style=taskExcerpt]
[TASK INPUT]
Task ID: mte.kyriba.cash-review.3k0802
Request: Hanover Re (Bermuda) USD cash outlook,
  1 Oct to 30 Jan 2025. Explain the exceptions
  and record what you verified.
Source: cash_movements, counterparty records
State: CM-2025-1008 forecast INCLUDED,
  bank status REJECTED

[NATIVE INTERACTION]
business-system tools: queries, updates,
  confirmation of the applied change

[SHARED FILE INTERFACE]
bash: create, inspect, search, modify, or execute files
from the workspace root, with multi-line writes through
one quoted here-document
\end{lstlisting}

\subsubsection{Visual}

This example asks the policy to build a site brief and repair what a preview reveals.

\casetitle{Static website task}{Visual}
\begin{lstlisting}[style=taskExcerpt]
[TASK INPUT]
Task ID: mte.webdev.plumbing.2451
Brief: plumbing services site with a search field,
  a service list, and a contact block
Deliver: static tree with index.html and assets
Checks: layout, typography, color, whitespace

[NATIVE INTERACTION]
render the page and inspect a preview image
  at a narrow viewport

[SHARED FILE INTERFACE]
bash: create, inspect, search, modify, or execute files
from the workspace root, with multi-line writes through
one quoted here-document
\end{lstlisting}

\subsection{Filesystem-Memory Trajectory Examples}
\label{app:camgv2:trajectory}

One resolved episode per environment appears below, drawn from learner steps 1 and 50 of the training run and quoted at
the round numbers the logger recorded. Every episode carries at least one message replacement, and in every
case an action after the replacement uses what the policy wrote before it. The listings show the write, the
replacement, the read, and the action that depends on it.

\subsubsection{Shop}

This training episode runs 54 rounds and resolves with return 0.98. Session 1 confirms an upholstery
material. After the boundary the policy derives the later requirement from the note it wrote, and the
choice in session 2 follows that record rather than the session prompt.

\casetitle{A recorded attribute decides a later session}{Shop}
\begin{lstlisting}[style=taskExcerpt]
[TASK INPUT]
webshop:agentmemory_5, coordinated living-room furnishing, six sessions

[NATIVE INTERACTION]  session 1
r1  search["MAKLAINE Contemporary Cream Velvet Down Filled Modular Sectional Sofa"]
r2  click[b09jt3nb6d]
r3  bash: mkdir -p .agent_memory && cat > .agent_memory/session1_note.md <<'EOF'
    +Confirmed upholstery material: velvet
    EOF
r4  click[buy now]                                    reward +0.143

[FILE-MEMORY WRITE, then OBSERVATION]  r5
bash: cat > .agent_memory/CONTINUATION.md <<'EOF'
objective: SESSION_2_OF_6 - Select mood/hanging lighting. ...
    observation: Purchase recorded. The next shopping session is ready.
                 Task family: bundled_shopping  Progress: 1/6
                 the session prompt names no earlier attribute

[READ BACK]  r6-r10
r6  bash: rg --hidden -n '^Confirmed ' . 2>/dev/null | timeout 10000 cat
r7  bash: cat .agent_memory/CONTINUATION.md
r8  bash: timeout 10000 rg --hidden -n '^Confirmed ' .          empty
r9  bash: cat .agent_memory/session1_note.md
r10 bash: cat .agent_memory/CONTINUATION.md | grep -A5 "Session 1 complete"

[DERIVED REQUIREMENT]  r11
bash: echo "Velvet -> choose leather"
r13 bash: cat .agent_memory/session1_note.md                  reward -0.01
r15 bash: cat .agent_memory/CONTINUATION.md | grep "upholstery material"
r18 bash: cat .agent_memory/session1_note.md | grep "upholstery"

[REWRITE REJECTED, then ACCEPTED]  r20-r21
r20 bash: cat > .agent_memory/session2_note.md <<'EOF'
    +Confirmed selected product: OJM Foot Stools Footrest PU L...
    observation: Filesystem checkpoint was not accepted (checkpoint_not_changed).
                 Your write was byte-identical to the existing file; change at
                 least one field value before rewriting it.
r21 bash: mkdir -p .agent_memory && cat > .agent_memory/CONTINUATION.md <<'AGENT_MEMORY_EOF'
    objective: SESSION_2_OF_6 ...

[NATIVE INTERACTION]  session 2, r23-r25
r23 search["OJM Foot Stools Footrest PU Leather Footstool ..."]
r24 click[b09q12wybp]
r25 click[buy now]                                    reward +0.143

[NATIVE OUTCOME]
Sessions 3-6 continue the loop. r46 records the rug color, r53 the
curtain material, and r54 closes the sixth purchase. The episode
resolves with return 0.98.

\end{lstlisting}

\subsubsection{Coding}

This training episode runs 72 rounds and resolves with full return. The checkpoint records the function
under repair, the suspected condition, and the file location. The environment replaces the messages twice,
and after each replacement the policy reads the checkpoint and continues from the recorded line.

\casetitle{A repair checkpoint and a re-run test after two replacements}{Coding}
\begin{lstlisting}[style=taskExcerpt]
[TASK INPUT]
Repair the persistent repository in /testbed for the reported issue

[NATIVE INTERACTION]  r1-r16
r1  bash: ls /testbed && echo "---" && ls /testbed/astroid 2>/dev/null | head -40
r1  bash: grep -rn "get_module_files" /testbed/astroid/*.py ... | head -20
r16 bash: sed -n '490,530p' /testbed/astroid/modutils.py

[FILE-MEMORY WRITE, then CONTEXT REPLACEMENT]  r17
bash: cat > .agent_memory/CONTINUATION.md <<'AGENT_MEMORY_EOF'
objective: Fix get_module_files() in /testbed/astroid/modutils.py which
returns an empty list when scanning directories.
evidence:
 - get_module_files at line 442, has a bug on the line that uses
   if not list_all or {"__init__.py", "__init__.pyi"}.issubset(filenames):
   dirnames[:] = ()
 - bug locati...                                          replace_messages

[READ BACK]  r18-r19
r18 bash: cat .agent_memory/CONTINUATION.md
r19 bash: sed -n '450,475p' astroid/modutils.py

[REJECTED WRITE, then REPLACEMENT]  r37-r38
r37 bash: cd /testbed && python3 -c "import tempfile, os
    from astroid import modutils ..."                     reward -0.1
    observation: Filesystem checkpoint was not accepted (checkpoint_not_changed).
r38 bash: cat > .agent_memory/CONTINUATION.md <<'AGENT_MEMORY_EOF'
    objective: Fix get_module_files() in /testbed/astroid/modutils.py ...
                                                          replace_messages
r39 bash: cat .agent_memory/CONTINUATION.md

[NATIVE OUTCOME]  r57
The policy fixes the condition and reports that all tests pass, with the
file, the line, and the failing condition named. Return 1.0.

\end{lstlisting}

\subsubsection{AutoResearch}

This training episode runs 34 rounds and resolves above full return. The checkpoint carries the task
identifier, the metric direction, the row counts, and the state of the validation, and the environment
replaces the messages twice.

\casetitle{A pipeline checkpoint rewritten after validation}{AutoResearch}
\begin{lstlisting}[style=taskExcerpt]
[TASK INPUT]
Task ID: nvidia-stock-historical-data@2
Native metric: NVDAIntradayRangeMetrics (lower is better)

[NATIVE INTERACTION]  r1-r6
r1  bash: cat TASK.md; cat data/description.txt; head -5 data/train.csv;
    head -5 data/test.csv; wc -l data/train.csv data/test.csv ...
r2  bash: python -c "import pandas as pd, numpy as np
    tr = pd.read_csv('data/train.csv'); te = pd.read_csv('data/test.csv') ..."

[FILE-MEMORY WRITE, then CONTEXT REPLACEMENT]  r7
bash: mkdir -p .agent_memory && cat > .agent_memory/CONTINUATION.md <<'AGENT_MEMORY_EOF'
objective=Predict NVDA intraday range metric (NVDAIntradayRangeMetrics,
lower is better) for 117 test rows (2023-12-29 to 2024-06-17) from 6275
training rows. Output submission.csv with columns row_id,target.
measured_validation_or_failure=validation: not measured yet.
                                                          replace_messages
r8  bash: cat .agent_memory/CONTINUATION.md

[REWRITE AFTER VALIDATION, then REPLACEMENT]  r23
bash: mkdir -p .agent_memory && cat > .agent_memory/CONTINUATION.md <<'AGENT_MEMORY_EOF'
objective=Predict NVDA intraday range metric ...           replace_messages
r24 bash: cat .agent_memory/CONTINUATION.md

[NATIVE OUTCOME]  r31
bash: submit
The submission is scored at 0.969 on the native metric, and the episode
resolves with return 1.02.

\end{lstlisting}

\subsubsection{General}

This training episode runs 33 rounds and resolves with return 0.6. The environment replaces the active
messages in r8-r9 and again in r17. Between them the policy recovers its checkpoint and goes on to query
the business system with the identifiers it recorded before the boundary.

\casetitle{A checkpoint that carries case identifiers across two replacements}{General}
\begin{lstlisting}[style=taskExcerpt]
[TASK INPUT]
Epic Games legal and finance quarterly review, September 30 2025:
reconcile the risk system against the case-progress workbook

[NATIVE INTERACTION]  r1-r7
r1  bash: ls -la && echo "---" && cat .agent_memory/CONTINUATION.md 2>/dev/null
    || echo "no memory file"
r3  bash: python3 -c "import docx; doc = docx.Document(
    'Epic_Google_Q3_2025_Litigation_Risk_Memorandum.docx') ..."
r7  bash: python3 -c "import openpyxl; wb = openpyxl.load_workbook(
    'Epic_Google_Q3_2025_Case_Progress_and_Fee_Ledger.xlsx') ..."

[FILE-MEMORY WRITE, then CONTEXT REPLACEMENT]  r8-r9
r8  bash: cat > .agent_memory/CONTINUATION.md <<'EOF'
    # EPIC-GOOGLE Q3 2025 RECONCILIATION - CONTINUATION
    ## Objective
    Create a mana...                                    replace_messages
r9  bash: mkdir -p .agent_memory && cat > .agent_memory/CONTINUATION.md <<'AGENT_MEMORY_EOF'
    objective: Reconcile ServiceNow IRM for...          replace_messages

[READ BACK, then TOOL QUERIES]  r10-r19
r10 bash: cat .agent_memory/CONTINUATION.md
r12 bash: python3 -c "import openpyxl; wb = openpyxl.load_workbook(
    'Epic_Google_Q3_2025_Case_Progress_and_Fee_Ledger.xlsx') ..."
r13 mcp__servicenow_irm__search_risk_matters(member_case_number=3:20-cv-05671-JD)
r15 mcp__servicenow_irm__list_exposure_forecasts(
        risk_matter_id=RSK-EPGGL-2025Q3-CUR, period_start=2025-01-01)
r17 bash: cat > .agent_memory/CONTINUATION.md <<'EOF' ...    replace_messages
r18 bash: cat .agent_memory/CONTINUATION.md
r19 mcp__servicenow_irm__calculate_forward_exposure(
        matter_id=MAT-EPIC-GOOGLE-2020-05671, forecast_period_start=2025-...)

[VERIFICATION]  r23
The policy states that the relevant data is verified and retrieves the
prior-quarter stored summary to confirm the stale-summary gap:
exposure_summary_usd_cents = 312640000, prior_overdue_control_count = 1

[NATIVE OUTCOME]
The episode resolves in r33 with return 0.6.

\end{lstlisting}

\subsubsection{Visual}

This training episode runs 76 rounds and resolves. The environment replaces the messages four times, and
after each replacement the policy reads its snapshot and resumes the build. The finding that decides the
final revision is a rendered preview, so the policy has to write down what it saw.

\casetitle{A site snapshot rewritten at every replacement}{Visual}
\begin{lstlisting}[style=taskExcerpt]
[TASK INPUT]
PROJECT BRIEF - EcoStay Rooms: a static marketing landing page for an
eco-friendly hotel-booking platform, delivered under dist/ with
dist/index.html as the entry point, no backend

[NATIVE INTERACTION]  r1-r2
r1  bash: mkdir -p dist/assets && ls -la /workspace
r2  the policy writes one index.html with embedded CSS and JS and inline
    SVG assets, so no external request is needed

[FILE-MEMORY WRITE, then CONTEXT REPLACEMENT]  r3
bash: mkdir -p .agent_memory && cat > .agent_memory/CONTINUATION.md <<'EOF'
# EcoStay Rooms - Continuation Snapshot
## Objective
Build a static marketing landing page for "EcoStay Rooms" ...
## Status
- Workspace: ...                                        replace_messages
r4  bash: cat .agent_memory/CONTINUATION.md

[SECOND WRITE, then REPLACEMENT]  r8-r9
r8  bash: cat > /workspace/.agent_memory/CONTINUATION.md <<'EOF' ...
                                                        replace_messages
r9  bash: cat .agent_memory/CONTINUATION.md

[THIRD AND FOURTH REPLACEMENTS]  r22-r23 and r46-r47
Each pair rewrites the snapshot with the sections already delivered and
reads it back, so the build resumes at the next section.

[NATIVE OUTCOME]  r64
The policy reports that the page is delivered under dist/index.html with
all eleven required sections present. Return 0.35.

\end{lstlisting}

\subsubsection{A memory decision from write to use}
\label{app:memory-decision}

Consider the Visual plumbing-site task in Appendix~\ref{app:camgv2:examples}.
Its brief specifies a responsive, static site with local search. During development,
a narrow-viewport preview may reveal that the plumber cards overflow. The site
files preserve the current implementation, but do not by themselves record which
preview exposed the defect, the proposed explanation, or whether a repair has been
checked. These are the memory decisions on which OPD must provide useful targets.

\paragraph{Scoring the written content.}
The student can keep the detailed inspection and the pending-work summary in
separate files. The following abbreviated actions show their different roles:

\noindent\begin{minipage}{\linewidth}
\begin{lstlisting}[style=memoryExample]
mkdir -p .agent_memory
cat > .agent_memory/design-review.md <<'EOF'
Narrow preview: plumber cards overflow.
Hypothesis: card min-width or grid columns.
Repair has not been visually verified.
EOF
cat > .agent_memory/CONTINUATION.md <<'EOF'
Local search is implemented on mock data.
Mobile card overflow remains unresolved.
Read design-review.md, inspect CSS, re-render.
EOF
\end{lstlisting}
\end{minipage}

At the continuation write, the teacher and the student see the earlier preview and inspection
observations. A content-target probe contrasts this record with one claiming that the
mobile layout was checked. Both writes are executable. Only the first preserves the
unresolved defect. Token-level teacher scores distinguish them in the file content
written through Bash. We compare that preference with subsequent native task outcomes rather
than assume that higher likelihood implies better memory.

\paragraph{Content-target probes.}
\label{app:content-target-probes}
At a fixed invocation, candidate writes retain or omit the same task-critical observation or
requirement. The candidates have the same executable structure, matched content-token length,
and identical information before the changed span. We compare teacher log likelihood per
written-content token under each candidate's own prefix, excluding shared command syntax from
the content score. Functional retention labels come from the task specification or a blinded
content annotation, not from the teacher's preference. Appendix~\ref{app:teacher-preference} reports the
measurement.

\paragraph{Recovering state after replacement.}
After the accepted continuation write, the wrapper removes the earlier interaction
history and preserves both files. It supplies the continuation-file path without
inserting either file's contents. The student's next recovery can be:

\noindent\begin{minipage}{\linewidth}
\begin{lstlisting}[style=memoryExample]
cat .agent_memory/CONTINUATION.md
cat .agent_memory/design-review.md
grep -nE 'min-width|grid-template' site.css
\end{lstlisting}
\end{minipage}
The first two reads recover the unresolved defect and its diagnostic hypothesis.
The third locates the implementation to inspect. The policy then edits the relevant
CSS and requests a new preview. Whether the repair worked is decided by the new image
and by the native grading of the final site.

\paragraph{What the case tests.}
This walkthrough links a teacher-scored write to a memory-dependent action. Detailed
notes and bounded continuations coexist. Neither must compress the other. A trained-policy
case records the task, write/read invocations, reset, post-read action, new verification,
and native outcome. A matched post-reset fork removes the defect description while
preserving website files to test its effect on recovery. The paired content intervention measures the effect of the defect
description, and this walkthrough shows the interface that carries it.

\subsection{Code Availability}
\label{app:camgv2:code}

Code will be released with the camera-ready version.

\section{A Cost Model for Interface Supervision}
\label{app:cost-model}

\subsection{The cost law}
\label{sub:cost-model}

Teaching one action takes $N\propto 1/(Qu)$ supervised occurrences while the student rarely emits it, where
$Q$ is the probability the student emits it and $u$ the teacher's preference for it. The derivation starts from
two published facts. The per-token reverse-KL logit gradient has $\ell_1$ norm equal to twice the
teacher--student log-probability gap times a student-side factor that grows as the
emitted token becomes less likely under the student~\cite{token-level-analysis}.
And the student's deficiency on the teacher's support splits into misallocation
plus a coverage term $-\log Q$, where $Q=\sum_{v\in\mathcal{K}}p_\theta(v\mid s)$ is
the student's mass on the teacher's top-$K$ set $\mathcal{K}$, with the coverage term invisible
to the sampled advantage~\citep{mismatch-matters}.

Definition~\ref{def:two-token} fixes the reduction, Lemma~\ref{lem:movement} gives
the movement per supervised occurrence, Proposition~\ref{prop:drift} combines the two
sampling outcomes into a drift, Theorem~\ref{thm:cost} integrates the drift into
the cost, Corollary~\ref{cor:two-factor} reads it as the two-factor law, and
Remark~\ref{rem:support} extends the form from one token to the teacher's top-$K$ support.

\begin{definition}[Two-token reduction]
\label{def:two-token}
Fix one prefix $s$ and reduce it to two tokens, $u$ preferred by the teacher and $y$
not. Write $z=\log\frac{p_\theta(u\mid s)}{p_\theta(y\mid s)}$ for the student's
log-odds and $z^{*}$ for the teacher's, so $Q=\sigma(z)=p_\theta(u\mid s)$ with
initial value $Q_0$ and target $Q^{*}=\sigma(z^{*})$, and let
\begin{equation}
D=\log\frac{q(u\mid s)}{p_\theta(u\mid s)},\qquad
E=\log\frac{p_\theta(y\mid s)}{q(y\mid s)} ,
\end{equation}
so that $D$ is the teacher--student gap on the preferred token, $E$ its magnitude on
the other, and $D+E=z^{*}-z$ at every point. Write $u_0=D_0+E_0$ for the initial
value of the gap. Let $\lambda$ be the effective logit step per supervised
occurrence, let $N$ count supervised occurrences of the prefix's tokens, not
optimizer steps, and call the token learned once $z^{*}-z$ has closed to a fraction
$\rho$ of $u_0$.
\end{definition}

The derivation works at the level of this prefix's logits, not its parameters:
what the parameters couple across prefixes, together with repeated supervision
within a round and the optimizer's normalized step, is absorbed into $\lambda$,
which is not the learning rate. The result is a proportionality: it fixes which
quantities sit in the numerator and which in the denominator, and $\lambda$ only
scales it.

\begin{lemma}[Per-occurrence movement]
\label{lem:movement}
One supervised occurrence of a token $x$ moves the log-odds $z$ by
$2\lambda A^{\mathrm{OPD}}(s,x)\big(1-p_\theta(x\mid s)\big)$.
\end{lemma}
\begin{proof}
This is the softmax identity of~\citet{token-level-analysis} in the two-token
reduction, where $z$ is the difference of the two logits and the update touches
only those logits.
\end{proof}

Which token the student samples decides the sign of the movement. Proposition~\ref{prop:drift} averages the two cases into a contraction on the gap.

\begin{proposition}[Both outcomes teach]
\label{prop:drift}
Sampling $u$ gives $\Delta z=2\lambda D(1-Q)$. Sampling $y$ gives
$\Delta z=2\lambda EQ$, because the score of $y$ is negative and its own mass is
pushed onto $u$. Both outcomes move $z$ toward $z^{*}$, and
\begin{equation}
\mathbb{E}\big[\Delta D\big]=-\,2\lambda\,Q\,(1-Q)^{2}\,\big(z^{*}-z\big)+O(\lambda^{2}),
\label{eq:drift}
\end{equation}
a rate of $2\lambda Q(1-Q)^2$ times the full log-odds gap. The relation is first order in the effective
logit step, which is the drift the continuum limit below integrates.
\end{proposition}
\begin{proof}
The $Q$ factor on the second outcome appears together with the $1-Q$ frequency of
that outcome, so the two contributions are of the same order in $Q$, not of the same
order in $D$. Using $\mathrm{d}D/\mathrm{d}z=-(1-Q)$ and
$\Delta z=2\lambda(1-Q)$ times the score of whichever token was sampled
(Lemma~\ref{lem:movement}) gives \eqref{eq:drift}. Expanding $D(z+\Delta z)$ about $z$ leaves a
second-order term in $\lambda$, which is what the $O(\lambda^{2})$ records and what vanishes when the
recursion becomes the differential equation below. Sampling $u$ alone would have given
$2\lambda Q(1-Q)^2D$, which is slower by the factor $(D+E)/D$.
\end{proof}

\begin{theorem}[Cost]
\label{thm:cost}
Under Definition~\ref{def:two-token}, write $u=z^{*}-z$ for the gap and let $N$ count the supervised
occurrences that close it to the fraction $\rho$. Separating variables in the drift \eqref{eq:drift} gives
the exact mean-field cost
\begin{equation}
N=\frac{1}{2\lambda}\int_{\rho u_0}^{u_0}\frac{\mathrm{d}v}{v\,Q(v)\,(1-Q(v))},\qquad Q(v)=\sigma(z^{*}-v),
\label{eq:exact-integral}
\end{equation}
from which three statements follow in their own ranges.

While the coverage stays at or below one half along the trajectory the drift gives the bound
\begin{equation}
N\;\le\;\frac{\log(1/\rho)}{\lambda\,Q_0},
\label{eq:bar}
\end{equation}
which uses only the initial coverage. In the rare-action regime, where the student rarely emits the token
throughout, so that $Q(v)\ll1$ and $1-Q(v)\approx1$ on the whole interval,
\begin{equation}
N=\frac{1+o(1)}{2\lambda\,Q_0\,u_0},\qquad Q_0\to0 \text{ with } Q^{*} \text{ and } \rho \text{ fixed},
\label{eq:two-factor}
\end{equation}
and the gap enters as a factor. When the coverage instead stays near a constant $\bar Q$, which is the
regime of a student that already emits the action, the same drift integrates to
\begin{equation}
N\;\approx\;\frac{\log(1/\rho)}{2\lambda\,\bar Q\,(1-\bar Q)},
\label{eq:near-full}
\end{equation}
and $u_0$ enters only through the starting point, not as a factor. The two regimes are limits of
\eqref{eq:exact-integral}, not extensions of one another: the second holds for a policy already close to
the teacher, where $\bar Q\approx Q^{*}$.
\end{theorem}
\begin{proof}
Read \eqref{eq:drift} as a recursion in the log-odds and pass to the continuum: $N$ draws move $z$ by
\begin{equation}
\frac{\mathrm{d}z}{\mathrm{d}n}=2\lambda\,Q(1-Q)\,(z^{*}-z),\qquad Q=\sigma(z).
\label{eq:ode}
\end{equation}
With $u=z^{*}-z$ this is $\mathrm{d}u/\mathrm{d}n=-2\lambda Q(1-Q)u$, and separating variables gives
\eqref{eq:exact-integral} directly, without any approximation on $Q$.

For \eqref{eq:bar}, $Q(1-Q)\ge Q/2$ while $Q\le\tfrac12$, and $Q$ is non-decreasing along the trajectory,
so $\mathrm{d}z/\mathrm{d}n\ge\lambda Q_0(z^{*}-z)$, which integrates to the stated bound.

For \eqref{eq:two-factor}, the sigmoid identity
$\big[Q(v)(1-Q(v))\big]^{-1}=e^{v-z^{*}}+2+e^{z^{*}-v}$ turns \eqref{eq:exact-integral} into
\begin{equation*}
N=\frac{1}{2\lambda}\int_{\rho u_0}^{u_0}\frac{e^{v-z^{*}}+2+e^{z^{*}-v}}{v}\,\mathrm{d}v ,
\end{equation*}
and the outer two terms stay bounded as $Q_0\to0$ while the first grows, so
\begin{equation*}
N=\frac{e^{-z^{*}}}{2\lambda}\int_{\rho u_0}^{u_0}\frac{e^{v}}{v}\,\mathrm{d}v\,(1+o(1)).
\end{equation*} With $Q^{*}$ held fixed and
$Q_0\to0$, $u_0\to\infty$ and the integral is carried by its upper limit, the exponential integral
satisfying $\mathrm{Ei}(u_0)-\mathrm{Ei}(\rho u_0)=\frac{e^{u_0}}{u_0}(1+o(1))$. Hence
\begin{equation*}
N=\frac{e^{u_0-z^{*}}}{2\lambda u_0}(1+o(1))=\frac{1-Q_0}{2\lambda Q_0u_0}(1+o(1)),
\end{equation*}
because $e^{u_0-z^{*}}=(1-Q_0)/Q_0$ exactly, the odds ratio of the two initial probabilities. Dropping the
$1-Q_0$ factor, which tends to one, leaves \eqref{eq:two-factor}.

For \eqref{eq:near-full}, $Q(v)\to\bar Q$ uniformly on $[\rho u_0,u_0]$ as the starting point approaches
the teacher, so the integrand of \eqref{eq:exact-integral} is $\big[2\lambda\bar Q(1-\bar Q)v\big]^{-1}$
to leading order and the integral is $\log(1/\rho)$ over that factor.
\end{proof}

\begin{corollary}[Two factors in the rare-action regime]
\label{cor:two-factor}
In the branch \eqref{eq:two-factor}, the cost has \emph{two} factors, and they are the two axes the
interface shares. Coverage $Q_0$ is how seldom the student emits the action on its own. The initial gap
$u_0=\operatorname{logit}(Q^{*})-\operatorname{logit}(Q_0)$ is how far the student's distribution starts from the teacher's, which sets
the size of the correction the supervision has to deliver. Once the student emits the action often enough
for that branch to end, \eqref{eq:near-full} governs the rest of the way and the gap stops multiplying the
cost.

The gap is a discrepancy between the two policies, not a measure of how strongly the teacher prefers the
action. A teacher at $Q^{*}=0.51$ against a student at $Q_0=0.001$ has $u_0\approx6.95$ while barely
distinguishing the two tokens, and a teacher at $Q^{*}=0.99$ against a student at $Q_0=0.98$ has
$u_0\approx0.70$ while preferring one of them decisively. For a fixed relative contraction, a smaller gap
means both a weaker push and less distance to cover, which is why $u_0$ leaves the cost altogether in
\eqref{eq:near-full}. A gap of the other sign, $Q^{*}<Q_0$, reverses the drift in \eqref{eq:drift} and moves
the student away from $u$. Whether the teacher prefers the right action is a separate question, which the
two-token dynamics do not settle and Appendix~\ref{app:teacher-preference} measures directly.
\end{corollary}

\begin{remark}[Numerical check]
\label{rem:numerics}
A direct simulation of the two-token recursion reproduces the constant in
\eqref{eq:two-factor} to within $30\%$ across three orders of magnitude in $Q_0$, for $u_0>0$. A
separate run with $u_0<0$, where \eqref{eq:two-factor} is not claimed since its right-hand side would be
negative, confirms the reversed direction of the drift. Replacing the single token by a support of
two tokens out of six reproduces the same scaling, with the ratio of measured to
predicted draws between three and four over a fortyfold range of $Q_0$. The constant is of
order one and grows with the size of the support.

Table~\ref{tab:regimes} evaluates \eqref{eq:exact-integral} and the rare-action form \eqref{eq:two-factor}
side by side. The two agree where the regime hypothesis holds, which is a small student coverage under a
teacher that is confident, and they part company once the teacher is nearly as rare as the student or the
required relative contraction is small, exactly the cases the theorem excludes.
\end{remark}

\begin{table}[t]
\centering
\small
\setlength{\tabcolsep}{4pt}
\begin{tabular}{@{}ccccc@{}}
\toprule
$Q_0$ & $Q^{*}$ & $\rho$ & \eqref{eq:exact-integral} & \eqref{eq:two-factor} \\
\midrule
$0.001$ & $0.5$ & $0.5$ & $82.7$ & $72.4$ \\
$10^{-5}$ & $0.5$ & $0.5$ & $4780.0$ & $4342.9$ \\
$0.001$ & $0.0012$ & $0.5$ & $329.9$ & $2739.4$ \\
$0.001$ & $0.5$ & $0.95$ & $21.7$ & $72.4$ \\
\bottomrule
\end{tabular}
\caption{\textbf{Where the rare-action form holds.} The two columns are $\lambda N$ from the exact integral and
from the asymptotic form, for a student that rarely emits the token under a confident teacher, under a teacher
nearly as rare as the student, and under a small relative contraction.}
\label{tab:regimes}
\end{table}

The cost law of Sec.~\ref{app:interface} is stated at the level of the teacher's
top-$K$ support, not a single token.

\begin{remark}[Extension to the teacher's support]
\label{rem:support}
Replacing the single token by the set $\mathcal{K}$ preserves the two-factor form: $D$ becomes the
deficit score of~\cite{mismatch-matters}, whose coverage term $-\log Q$ responds to mass entering
$\mathcal{K}$ exactly as $-\log Q$ did here, both derivatives being $-(1-Q)$ with respect to the logit
pushed into $Q$, and the misallocation term inside $\mathcal{K}$ contracts by the argument of
Proposition~\ref{prop:drift} applied to the renormalized distributions. The simulation of
Remark~\ref{rem:numerics} confirms that only the constant changes.

This is an extension of the form, not a proof of it at arbitrary support. Split the vocabulary into
$\mathcal{K}$ and its complement and the reverse KL separates into a Bernoulli term in the mass $Q$ plus two
conditional terms,
$D_{\mathrm{KL}}(p\|q)=D_{\mathrm{KL}}\big(\mathrm{Bern}(Q)\|\mathrm{Bern}(Q^{*})\big)
+Q\,D_{\mathrm{KL}}(p_{\mathcal{K}}\|q_{\mathcal{K}})
+(1-Q)\,D_{\mathrm{KL}}(p_{\bar{\mathcal{K}}}\|q_{\bar{\mathcal{K}}})$,
and the conditional terms let mass move inside the support without following the single-token dynamics we
integrated. The two-token cost law applies to $\mathcal{K}$ as a whole with the same two factors and an
order-one constant; a claim that it holds verbatim for every $K$ would need that separation carried through,
which we do not do here.
\end{remark}

Section~\ref{sec:results} reports the pre-training cost of supervision on each
interface. The algebra fixes what that cost depends on.

\subsection{The interface as a design variable}
\label{app:interface}

The objective sets the estimator and the teacher is frozen. The one design variable left
is the interface the supervision passes through. Supervision arrives only
inside an intersection: the action must be one the student emits on its own and one whose
consequence the teacher can judge. Outside it the failures differ in kind
(Table~\ref{tab:interface-cells}): an action the student never emits (when to call, which key
to use) or an emitted action the teacher scores uninformatively (argument order, a key
schema). An action the student never emits still moves under the drift of
Proposition~\ref{prop:drift}, since a sampled alternative pushes probability toward it, but the
teacher then never supervises that action's own structure or content. The state axis has already answered the same question. OPD is valid only when the teacher scores
the state the student actually acted in, and reconstructing that state is what makes
compact-memory distillation work~\citep{liu2026memopd}. The interface axis is the one still
open.

\begin{table}[!t]
\centering
\small
\setlength{\tabcolsep}{3pt}
\renewcommand{\arraystretch}{1.06}
\begin{tabular}{@{}>{\raggedright\arraybackslash}p{0.20\linewidth}>{\raggedright\arraybackslash}p{0.37\linewidth}>{\raggedright\arraybackslash}p{0.35\linewidth}@{}}
\toprule
 & \emph{teacher prefers it} & \emph{teacher does not} \\
\midrule
\emph{student emits it} & supervision: the intersection & excess: negative advantage, growing with the gap \\
\emph{student does not} & deficit: no direct supervision of the ungenerated action & empty: absent from the loss \\
\bottomrule
\end{tabular}
\caption{\textbf{Interface supervision has four cells.} Rows split by whether the student
emits the action, columns by whether the teacher prefers it. Only the intersection provides direct
supervision on the desired action and its content. Eq.~\ref{eq:cost} prices it.}
\label{tab:interface-cells}
\end{table}

What the intersection constrains is the action, not what is written. What to retain and what to recover, and when to do either, can all be behavior neither model
has. The intersection is necessary but not sufficient: the
chosen action must also leave state that persists beyond the active context and can be read
back, which is why we do not supervise ordinary turns of dialogue.

In the rare-action, large-discrepancy regime of Theorem~\ref{thm:cost}, teaching one action takes
\begin{equation}
\underbrace{N}_{\substack{\text{updates to}\\\text{learn it}}}
\;\propto\;
\frac{1}{\underbrace{Q_0}_{\text{student emits it}}\;\;
      \underbrace{u_0}_{\text{initial gap}}}.
\label{eq:cost}
\end{equation}
The deficit leaves the action itself unsupervised: the sampled advantage compares the two
distributions only where the student went~\citep{mismatch-matters}, so an action the student never emits
receives no direct supervision of its own structure or content, only the indirect movement that
Proposition~\ref{prop:drift} gives it. The excess is destructive: an action the teacher
scores below the student's own probability receives a negative advantage that grows as the
gap widens. The probability it frees returns to actions the student already prefers rather
than to the one the teacher would have chosen~\citep{asymmetric-opd,mismatch-matters}. The empty cell never enters the
loss, so the interface is never learned. A weak or wrong preference has two causes: a
next-token score is myopic about the outcome even for a capable
teacher~\citep{myopic-scorer,when-to-follow} — hence token-level reweighting
(Sec.~\ref{sec:related}) — and unfamiliarity, the one an interface can remove.

Dedicated memory tools fall outside the intersection twice over. A protocol call is a handful of tokens in which one error forfeits the entire action, so every teacher error is expensive. Its correctness also
follows a convention the visible context does not reveal, which is exactly what a general
teacher can be wrong about~\citep{harne2026privileged}, so teacher errors are frequent as
well. The weighting compounds it: reverse KL spends its early signal on the tail. The head is already
matched~\citep{kl-rethinking}, sampled-token objectives give improbable tokens disproportionate
gradient mass~\citep{token-level-analysis} (none under a renormalized top-$k$
view~\citep{topk-tail-grad}), and the teacher's least reliable positions receive the most
weight.

CAMP-II is built on this. Memory goes through actions the teacher and the student already
perform, so the teacher scores content it can judge. We transfer the
judgment of what to retain. What a harness computes across calls stays in the
harness~\citep{scaffold-separation}. The two models agree on the interface before training
starts, because we choose it.

\section{CAMP-II Training Details}
\label{app:distillation}
\newcommand{\HPUpdates}{100}
\newcommand{\HPEpisodes}{64}
\newcommand{\HPExposures}{6,400}
\newcommand{\HPDomainExposures}{1,280}
\newcommand{\HPSequence}{98,304}
\newcommand{\HPPrompt}{65,536}
\newcommand{\HPResponse}{32,768}

\brokenpenalty=10000

\subsection{Training Procedure}

Training is asynchronous: rollout replicas generate episodes with the published weights while the
learner continues to update, so an episode can enter an optimizer minibatch several weight
publications after it was sampled. Completed episodes enter a persistent queue, from which the
learner assembles minibatches without waiting for the slowest episode of a batch
(Algorithm~\ref{alg:camp-ii}). Each token retains the behavior log-probability recorded at
generation, the denominator of the importance ratio in Eq.~\ref{eq:opd-loss}. Weights are
published to the replicas after every learner round.

\noindent Algorithm~\ref{alg:camp-ii} lists the complete procedure, with the collection steps that the
rollout replicas run and the learning steps that the queue feeds.

\begin{algorithm}[t]
\caption{CAMP-II: file-memory OPD}
\label{alg:camp-ii}
\footnotesize
\begin{algorithmic}[1]
\Statex \textbf{Input:} initial student weights $\theta_0$, step size $\eta$, frozen teacher $\pi_T$, actions
$\{\mathcal A_d\}$, learner rounds $R$, episodes per round $N$, horizon $H$, episode queue $\mathcal Q$, memory interface $m$
\Statex \textbf{Output:} $\theta$
\Statex \textbf{Init:} $\theta\gets\theta_0$, $\mathcal Q\gets\emptyset$, $\theta^-\gets\theta$ for all rollout workers
\For{$r=1,\dots,R$} \Comment{one learner round}
  \For{each task $k$ assigned to a worker} \Comment{rollout}
    \State $h_1\gets$ initial context in a fresh episode-private workspace
    \For{$t=1,\dots,H$ until the episode terminates}
      \State $a_t\sim\pi_{\theta^-}(\cdot\mid h_t)$ \Comment{$\theta^-$: current worker weights}
      \State record $(x_i,s_i,\log\mu_i)$ for every token of $a_t$
      \State $o_t\gets$ observation the environment returns for $a_t$
      \State $h_{t+1}\gets$ the next visible context after $a_t$
      \If{the context is replaced}
        \State $h_{t+1}\gets$ workspace path of the continuation file
      \EndIf
    \EndFor
    \State $q_i\gets\pi_T(x_i\mid s_i)$ for every recorded token \Comment{frozen teacher}
    \State enqueue the episode with its $(x_i,s_i,\mu_i,q_i)$ and outcome
  \EndFor
  \State $\mathcal D\gets N$ episodes dequeued from $\mathcal Q$ \Comment{learner}
  \State $\theta\gets\theta-\eta\,\mathrm{AdamW}\big(\nabla_\theta\mathcal L^m_{\mathrm{OPD}}(\theta)\big)$
  \Comment{Eq.~\ref{eq:opd-loss}, clipped at $1\pm\epsilon$}
  \State publish $\theta$, and $\theta^-\gets\theta$ for all rollout workers
\EndFor
\end{algorithmic}
\end{algorithm}

\subsection{Training Configuration}

Table~\ref{tab:training-config} gives the shared configuration: model roles, the objective and its
clamps, the schedule with the per-environment observation budgets, the optimizer, batching, and
runtime, and records the single training seed. The teacher is frozen and receives
the same visible text and image observations as the student. The objective rows instantiate
Eq.~\ref{eq:opd-loss}, with the clamps and the dual-clip bound as numerical safeguards. A file
read is itself a sampled action, so the contents it returns are observations and a write is not
supervised a second time. There is no per-episode reweighting, so long episodes contribute
proportionally more tokens. The task reward is $R(\tau)=\sum_{t=1}^{T} r_t$, where $r_t$ is
nonzero only at the events of each environment's native reward contract
(Table~\ref{tab:reward-contract}). With optional task rewards, $\mathcal L_{\mathrm{PPO}}^m$ is
added to $\beta\mathcal L_{\mathrm{OPD}}^m$ using the same current-to-behavior student ratio.
The \HPUpdates{} learner rounds consume
\HPExposures{} episode exposures, \HPDomainExposures{} per domain under balanced interleaving.
These are exposures, not unique task counts. Failed native tasks remain training episodes, and
infrastructure retries retain their logical task identity and enter compute accounting.

\paragraph{Compute.} Training CAMP-II-4B took about 250 GPU-hours on eight GPUs, four training
the policy, two generating rollouts and two serving the frozen teacher, across the
\HPUpdates{} learner rounds.

\paragraph{Conditional objective.}
When task rewards are included, GAE runs over the concatenated sampled-token axis of each
trajectory~\citep{schulman2016gae}. A call's native reward is placed on the final token of its
response row, and preceding tokens receive zero immediate reward. At a row boundary the recursion
continues at the next row's first sampled-token state, including across context replacement.
Terminal task outcomes are counted once, and mechanical resets add no token or reward-bearing row.
For token position $n$, with $r_n$ the sparse reward and $d_n$ indicating trajectory termination,
and with the critic fixed during target computation,
\begin{align*}
\delta_n &= r_n + \gamma(1-d_n)V_{n+1}-V_n,\\
A_n &= \delta_n + \gamma\lambda(1-d_n)A_{n+1}.
\end{align*}
The return target is $G_n=A_n+V_n$ with zero bootstrap at native terminal and task-budget
termination. Task actor advantages are centred within each route, scaled by a shared factor, and
weighted by valid-token counts so routes carry equal weight; critic returns are unchanged. Each
actor or critic minibatch accumulates its own microbatches before its optimizer step. These task
advantages are distinct from the detached distillation signal $A_i^{\mathrm{OPD}}$. The pure-OPD
realization disables the task-reward loss and the learned critic, so task-return statistics do not
enter its actor gradient.

\begin{table*}[!t]
\centering
\scriptsize
\setlength{\tabcolsep}{3pt}
\renewcommand{\arraystretch}{1.06}
\begin{tabular}{@{}>{\raggedright\arraybackslash}p{0.19\textwidth}>{\raggedright\arraybackslash}p{0.43\textwidth}>{\raggedright\arraybackslash}p{\dimexpr0.38\textwidth-12pt\relax}@{}}
\toprule
Group & Hyperparameter & Value \\
\midrule
Models & Student (full parameters) & Qwen3.5-4B \\
 & Frozen external teacher & Qwen3.8-27B \\
\midrule
Objective & Signal & Sampled reverse KL ($k_1$) \\
 & Teacher target support & Student-sampled token \\
 & Target temperature & 1 \\
 & Supervised positions & Full sampled response \\
 & Sampled-token loss weight & 1 \\
 & Loss weight (inputs, observations, padding) & 0 \\
 & PPO clip & 0.2 \\
 & Negative-signal dual clip & 3 \\
 & Signal clamp & $\pm 10$ \\
 & Log-ratio clamp & $\pm 20$ \\
 & Task-PPO loss & Off \\
 & Learned critic & Off \\
 & Entropy coefficient & 0 \\
 & Auxiliary KL coefficient & 0 \\
\midrule
Schedule & Learner rounds & 100 \\
 & Episodes per learner round & 64 \\
 & Passes per episode batch & 1 \\
 & Domain mixture & Equal, balanced interleaving \\
 & Training seed & 1 \\
 & Data-loader seed & 4096 \\
\midrule
Optimizer & Optimizer & AdamW \\
 & Learning rate & $10^{-6}$ (constant) \\
 & AdamW $(\beta_1,\beta_2)$ & $(0.9,0.999)$ \\
 & AdamW $\epsilon$ & $10^{-8}$ \\
 & Weight decay & 0.01 \\
 & Warmup updates & 0 \\
 & Gradient norm limit & 1.0 \\
\midrule
Batching & Responses per GPU microbatch & $\leq 8$ \\
 & Tokens per GPU microbatch & $\leq 98,304$ \\
 & Gradient accumulation & One step per optimizer minibatch \\
 & Rows per optimizer minibatch & 512 \\
\midrule
Runtime & Precision & BF16 \\
 & Sharding & FSDP2 \\
 & Gradient checkpointing & On \\
\midrule
Sequence & Prompt + sampled response & 98,304 tokens \\
 & Prompt ceiling & 65,536 tokens \\
 & Response ceiling & 32,768 tokens \\
 & Route observation envelope & 8,192 (Shop, Coding, AutoResearch), 12,288 (General, Visual) tokens \\
\bottomrule
\end{tabular}
\caption{\textbf{Training configuration.}}
\label{tab:training-config}
\end{table*}

\begin{table*}[!t]
\centering
\small
\setlength{\tabcolsep}{3.5pt}
\renewcommand{\arraystretch}{1.08}
\begin{tabular}{@{}>{\raggedright\arraybackslash}p{0.115\textwidth}>{\raggedright\arraybackslash}p{0.84\textwidth}@{}}
\toprule
\textbf{Environment} & \textbf{Reward} \\
\midrule
Shop & $r_t=\tfrac{1}{7}$ per correct purchase of the session target under the cumulative budget; a
wrong purchase pays $-0.5$ and terminates the episode with earlier payments retained. Success requires
the full six-session sequence. \\
Coding & $R=\mathbf{1}[\text{hidden tests pass on the submitted repository change}]$ at the submission
sentinel; $r_t=-0.01$ per recoverable invalid action and at horizon failure. \\
AutoResearch & $R=\max(0,\,m)+0.05\,\mathbf{1}[\text{first execution}]+0.10\,\mathbf{1}[\text{first
validation}]$, where $m$ is the private grader's native-metric score of the submitted file and milestone
payments are capped at $0.15$ per episode; $r_t=-0.01$ at action-budget exhaustion or wall-clock
termination, and invalid submissions pay $0$. \\
General & $R=\mathbf{w}^{\top}\mathbf{r}$, a weighted rubric over the final answer, delivered files,
and post-action system state, paid once at finish with required rule checks as gates; $r_t=-0.1$ per
invalid action and $-0.01$ when the budget is exhausted without a submission. \\
Visual & $R=\mathrm{mean}(r_{\mathrm{rubric}},\,r_{\mathrm{fulfilment}},\,r_{\mathrm{assets}})$, an
absolute rubric over screenshots of the delivered page at finish (layout, typography, color,
whitespace, content), averaged with query fulfilment and asset quality; invalid actions pay $0$. \\
\bottomrule
\end{tabular}
\caption{\textbf{Task rewards.}}
\label{tab:reward-contract}
\end{table*}

Table~\ref{tab:training-config} also fixes the shared sequence budgets, extending the shared runtime
of CAMG~\citep{luo2026camg}. The budgets distinguish policy responses, environment calls, and
context replacement: an episode allows 64 policy responses, and a response containing up to
ten tool calls consumes one response budget unit. The calls execute in order and each
produces an observation. File writes and reads consume the same response budget as task
actions. A mechanical reset consumes no policy response, and Shop's six purchases share one
episode budget. Native command timeouts, evaluator limits, and task-artifact capacities
remain attached to the source task package.

\paragraph{Length accounting.}
The prompt ceiling is \HPPrompt{} tokens and the response ceiling is \HPResponse{} tokens,
within a \HPSequence{}-token sequence envelope. Declared observation envelopes are 8,192 tokens
for Shop, Coding, and AutoResearch, and 12,288 for General and Visual. Continuation pressure
uses a bounded ordinary-observation allowance of up to 4,096 tokens, or 5,120 on the larger-
observation routes, further limited by available headroom. The wrapper also reserves the
next response, control-message framing, a bounded continuation write, and one bounded retry.
It requests continuation
before that projection reaches the effective prompt capacity, provided enough actions remain
to write, read, and continue. The request point therefore depends on framing and retry
reserves, rather than a single fixed prompt threshold.
Counting uses the actual chat template and processor-expanded multimodal prompt. Visual
previews remain image inputs for both teacher and student. Base64 image payloads are never
tokenized as ordinary text. Candidate prompt lengths are checked before serving. Reserving
space alone does not guarantee that a returned observation fits. A response-length cap does
not automatically make a task an infrastructure failure.

If a nonterminal candidate exceeds a hard prompt or sequence ceiling, the wrapper terminates
the episode for context-budget exhaustion without serving the invalid prompt.
It counts as an unsuccessful task-budget termination, retains earned native rewards and
valid sampled actions, and is not retried or excluded as infrastructure failure. Already
terminal native outcomes are preserved. Budget termination has no future continuation value.
All interfaces follow the same rule.

\paragraph{Two file roles, one workspace.}
The 8,192-byte continuation limit applies only to the designated file written before context
replacement. Other memory files, such as task notes, remain freely organized in the persistent
workspace within its native capacity. The policy need not create these notes before writing the continuation file
or derive the continuation file from them. The wrapper validates the continuation write and clears active context. It supplies the
file path without injecting its contents. The subsequent policy-authored read consumes an
ordinary response. Interface controls preserve these budgets and state transitions, changing
only the declared memory interface.

\subsection{Across training seeds}

Table~\ref{tab:five-seeds} repeats the full CAMG-II evaluation for five training seeds of CAMP-II-4B. The
spread is small and the ordering of the domains is stable, so the comparison in the main text does not rest
on one draw.

\begin{table*}[t]
\centering
\small
\setlength{\tabcolsep}{8pt}
\begin{tabular*}{\textwidth}{@{\extracolsep{\fill}}lcccccc@{}}
\toprule
Seed & Shop & Coding & AutoResearch & General & Visual & Avg. \\
\midrule
1 & 100.0 & 37.5 & 64.1 & 81.2 & 40.2 & 64.6 \\
2 & 100.0 & 39.1 & 65.6 & 82.0 & 41.0 & 65.5 \\
3 & 100.0 & 38.3 & 63.3 & 80.5 & 39.7 & 64.4 \\
4 & 100.0 & 36.7 & 64.8 & 82.8 & 40.5 & 65.0 \\
5 & 100.0 & 38.3 & 66.4 & 81.2 & 40.0 & 65.2 \\
\midrule
Mean $\pm$ Std. & 100.0 $\pm$ 0.0 & 38.0 $\pm$ 0.9 & 64.8 $\pm$ 1.2 & 81.5 $\pm$ 0.9 & 40.3 $\pm$ 0.5 & 64.9 $\pm$ 0.5 \\
\bottomrule
\end{tabular*}
\caption{\textbf{Five training seeds of CAMP-II-4B.} Task success on the CAMG-II held-out panels, one row per
seed, with the mean and the standard deviation over seeds. Seed 1 is the run the main text reports. The
average column weights the five domains equally, and the Visual column follows the paper's definition of that
column.}
\label{tab:five-seeds}
\end{table*}

\subsection{Training dynamics}
\label{app:distillation:dynamics}

\begin{figure*}[t]
\centering
\includegraphics[width=\textwidth]{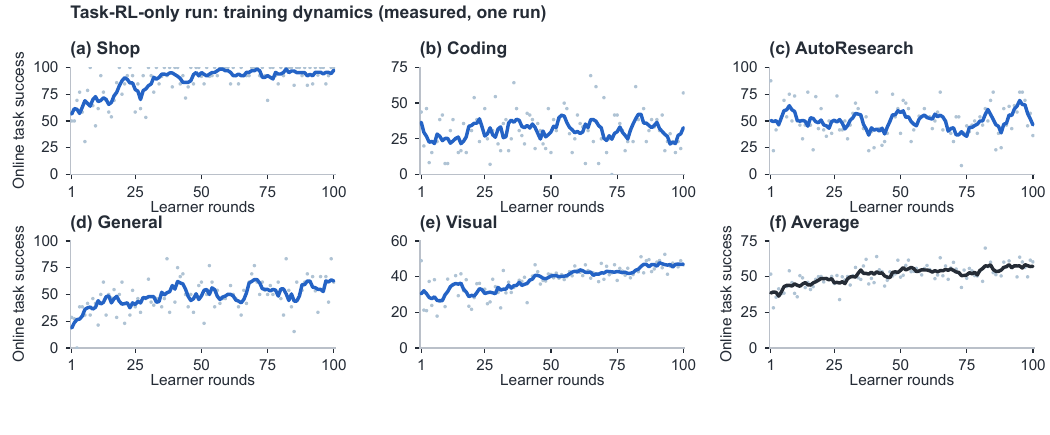}
\caption{\textbf{CAMP-I training dynamics.} The average in panel (f) weights the five domains equally.}
\label{fig:campii-training-curve}
\end{figure*}

\begin{figure*}[t]
\centering
\includegraphics[width=\textwidth]{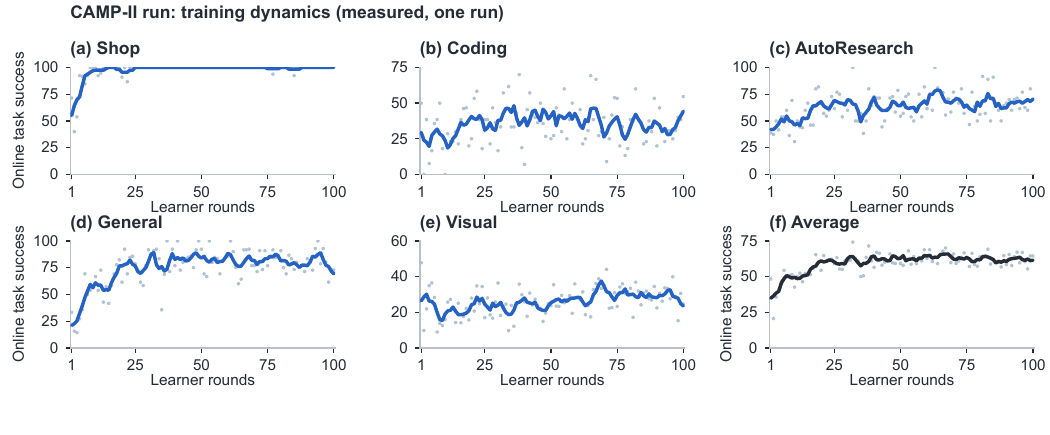}
\caption{\textbf{CAMP-II training dynamics.} Panel y-axes match
Fig.~\ref{fig:campii-training-curve} for a common scale. The average in (f) weights the five domains equally.}
\label{fig:campii-method-training-curve}
\end{figure*}

Figs.~\ref{fig:campii-training-curve} and~\ref{fig:campii-method-training-curve} report measured
online task success for CAMP-I (task rewards, no distillation teacher)
and CAMP-II (file-memory distillation teacher, no task rewards):
temperature-1 sampling on the training
task pools, \HPEpisodes{} episodes per learner round, one training run per configuration under one seed.
An update denotes one learner round, not a single optimizer-minibatch step.
Lines are five-update centered bucket means
and points are per-round values. Visual reports the mean judged grade rather than a binary
success rate. The remaining panels and the average use episode success. These
are training-pool measurements and are not comparable to the test-set evaluation of
Table~\ref{tab:variant-status}.

\section{Additional Experiments}
\label{sec:additional-experiments}

This appendix reports five measurements of the memory interface that the experiments in the main text leave
out. Three are taken before training, on how closely the teacher and the student already agree on memory
content, how much correction each interface asks of the student, and whether the teacher prefers the write that
carries the task's information. Two come from the trained policy, on whether the memory behaviour needs the
requirement to write at a boundary and on where the policy writes and reads its state around a replacement.

\subsection{Teacher--student agreement before training}
\label{app:content-tokens}

Before any training, the teacher and the student already organize memory content alike, and they do so most
closely under the file interface. To test whether an interface's notes fall inside what both models already
represent, we take the initial Qwen3.5-4B checkpoint and compare the teacher's and the student's similarity
structure over the notes each interface writes. Each interface writes its memory in one of three forms: a
file, a stored record for the memory tools, and a context summary for compaction. We call the text of any of them a
note. The student produces a note by receiving that interface's prompt with the call already opened and
completing the text, and each interface yields $160$ notes. We compare the two models on \emph{memory-content
tokens}, the tokens of the note body, leaving out the syntax that frames the write, the wrapper of the call and
the labels of its arguments. The frozen teacher (Qwen3.8-27B) and the student read the identical student-written sequence. For each
note we average the last-layer states of its content tokens into one vector, and from those vectors we build the teacher's and
the student's $160\times160$ cosine-similarity matrices over the notes. We compare the two matrices by the
rank correlation of their $12{,}720$ note pairs.

\begin{figure*}[!t]
  \centering
  \includegraphics[width=0.98\textwidth]{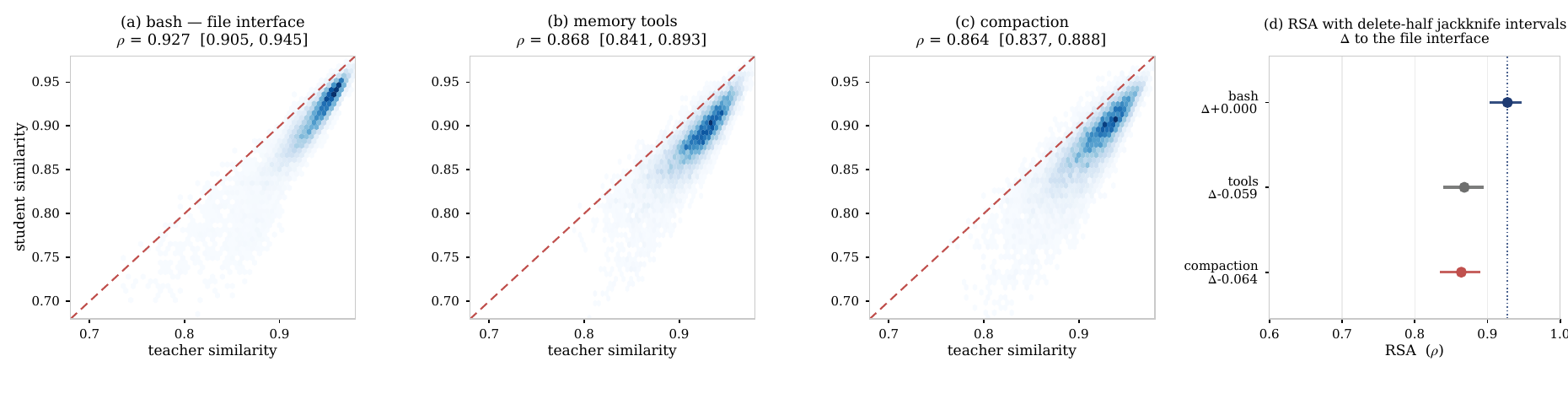}
  \caption{\textbf{The file interface keeps the teacher's organization of memory content.}
  Each interface writes its own note ($160$ notes per interface, initial Qwen3.5-4B). Panels (a)--(c) plot the
  similarity the teacher assigns to a note pair against the similarity the student assigns, over the
  $12{,}720$ pairs of each interface, so a pair on the dashed line is one the two checkpoints treat alike.
  Panel (d) gives the rank correlation of the two similarity matrices with delete-half jackknife intervals over the notes,
  highest for the file interface.}
  \label{fig:content-tokens}
\end{figure*}

The rank correlation is \textbf{0.927} for the file interface, \textbf{0.868} for the memory tools and
\textbf{0.864} for compaction (Fig.~\ref{fig:content-tokens}d), with delete-half jackknife intervals over the notes
$[0.905,0.945]$, $[0.841,0.893]$ and $[0.837,0.888]$. As a null control we shuffle the note identities of one matrix
before correlating, and the correlation falls below $0.09$ under all three interfaces. The file interface is therefore the
one whose representation of the notes the student already shares: the two checkpoints agree on which notes
resemble which, while under the memory tools and compaction the student organizes the same content more
loosely.

\subsection{Correction induced by each interface}
\begin{figure*}[!t]
  \centering
  \includegraphics[width=0.98\textwidth]{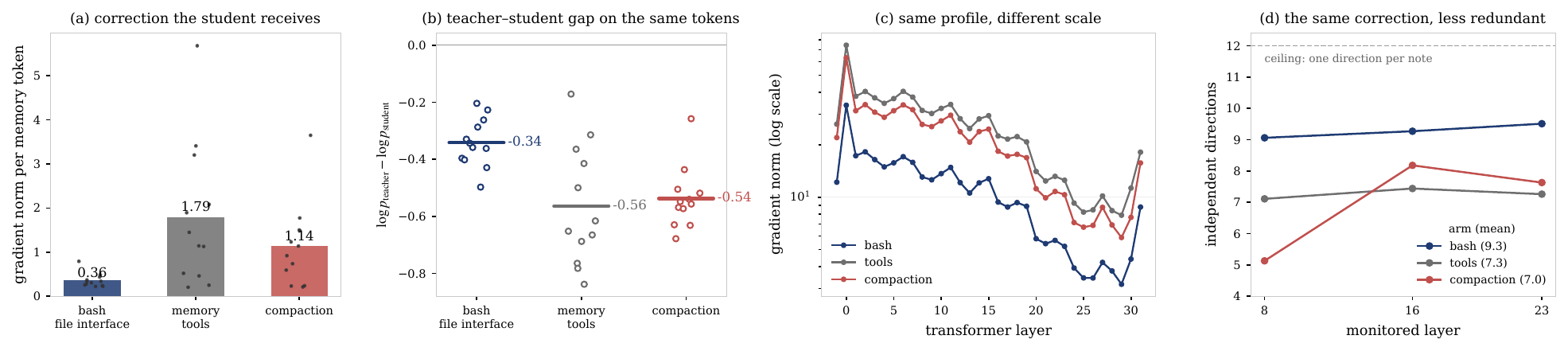}
  \caption{\textbf{The file interface generates the least corrective signal.}
  (a) Norm of the gradient of the distillation loss restricted to memory-content tokens, per supervised
  token, backpropagated through the initial checkpoint. Markers are individual notes ($n=12$ per interface)
  and the bar is their mean. (b) Teacher-minus-student log-probability gap on the same tokens, which is the
  loss weight. (c) Per-layer gradient norms: the three interfaces share one profile and differ in scale.
  (d) Nearly independent directions the twelve note-level gradients span, for the MLP gate projection of
  layers $8$, $16$ and $23$ (participation ratio of the Gram spectrum, where one direction per note is the
  ceiling of $12$): the file interface needs less correction per token, and that correction is less
  redundant.}
  \label{fig:interface-gradient}
\end{figure*}

The file interface asks for the least correction of the three, and that correction is the most independent. We measure the correction as the norm of the gradient the distillation loss
gives the parameters of the initial checkpoint when the loss is fitted to the notes of one interface, per
supervised token. This probe uses twelve notes per interface. Applying the loss of Eq.~\ref{eq:opd-loss} to those notes and restricting it to the
memory-content tokens, the gradient norm is $0.36$ for the file interface, $1.79$ for
the memory tools and $1.14$ for compaction (Fig.~\ref{fig:interface-gradient}a). The norm tracks the teacher--student probability gap on the same tokens
(Fig.~\ref{fig:interface-gradient}b), because the loss weights each token by that gap. On file-interface notes
the teacher's probability for the student's own tokens is close to the student's, so little correction remains,
while under the memory tools and compaction the gap is about $1.6$ times larger than under the file interface. Per layer the three interfaces follow one profile and differ only in scale
(Fig.~\ref{fig:interface-gradient}c). Note-level gradients can also repeat each other, and repeated directions teach the student little that is new,
so the probe measures a second quantity, how independent they are: for the MLP gate projection of
layers $8$, $16$ and $23$ we build the Gram matrix of the twelve note-level gradients and take the
participation ratio of its spectrum, which measures how many nearly independent directions the notes move the
student along, with one direction per note as the ceiling. The file interface reaches $9.3$ effective directions, the memory tools
$7.3$ and compaction $7.0$ (Fig.~\ref{fig:interface-gradient}d), so its note gradients span nearly independent directions, while those of the other two interfaces span fewer
independent ones. 

\subsection{Teacher preference over memory content}
\label{app:teacher-preference}

The teacher prefers the write that carries the task's own information, and it does so most often under the
file interface. We measure the preference at two granularities, with the whole note and with the segment of it
that carries the task's information. For each interface we first pair two notes written by the same writer
under that interface for different task contexts. Under either context's prefix the note written for that context carries the task's
information and the other does not, so the pair isolates whether the content matches the task. We score both
notes under that prefix with the frozen teacher, as the mean log-probability per memory-content token, and count
a decision as correct when the teacher ranks the matching note above the foreign one. A teacher that cannot tell
the two apart reaches $0.5$ by construction.

\begin{figure*}[!t]
  \centering
  \includegraphics[width=0.98\textwidth]{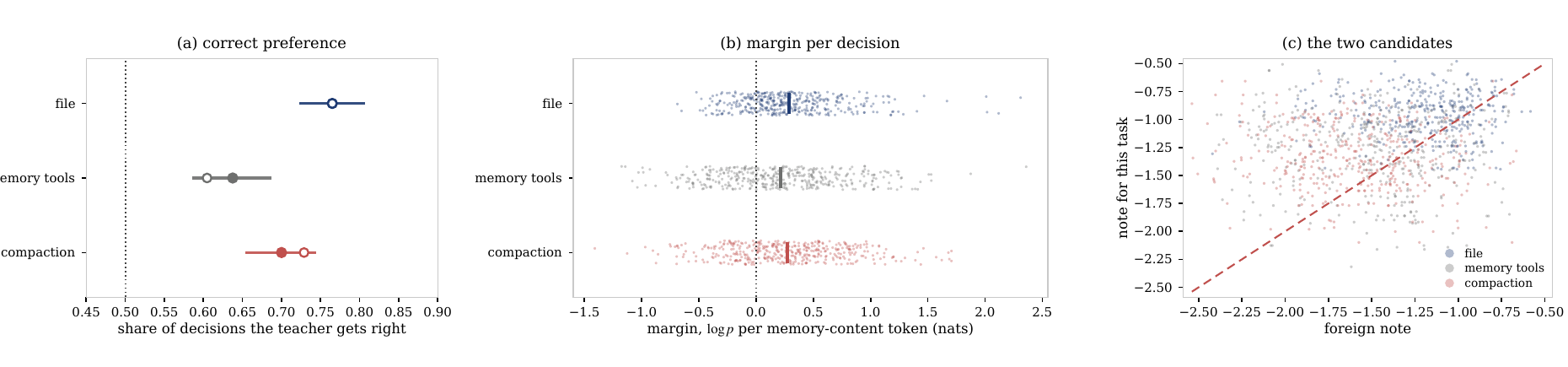}
  \caption{\textbf{The teacher ranks the note written for its own task above a foreign note, most often under
  the file interface.} (a) Share of decisions in which the teacher scores the note written for the prefix's own
  context higher, with intervals over decisions and the chance level of one half. Hollow markers restrict the
  decisions to pairs whose two notes differ by at most a fifth in content tokens. (b) Margin per decision, the
  difference in mean log-probability per memory-content token, markers are single decisions and the bar is their
  mean. (c) The two candidates' scores for the same decision, so a point above the diagonal is a decision the
  teacher gets right.}
  \label{fig:teacher-preference}
\end{figure*}

Over $200$ pairs per interface, which give $400$ decisions, the teacher is correct in \textbf{76.5\%} of
decisions for the file interface, \textbf{70.0\%} for compaction and \textbf{63.8\%} for the memory tools
(Fig.~\ref{fig:teacher-preference}a). The file interface's interval meets compaction's only over a narrow
band, $[0.720,0.745]$, just under a third of the width of either interval. The decisions are also not
independent, since every note is scored in both orders and each decision is made under one of the $47$
contexts. Comparing the interfaces over those contexts, the file interface leads the memory tools by $11.6$
points, with a paired interval of $[0.029,0.205]$ and $31$ of the $47$ contexts in its favour, and it leads
compaction by $5.9$ points, with $27$ contexts in its favour and $17$ against. The mean
margins are $+0.286$, $+0.271$ and $+0.212$ nats (Fig.~\ref{fig:teacher-preference}b). Restricting to decisions whose two notes differ by at most a
fifth in content tokens leaves the ordering unchanged, at $0.765$ for the file interface, $0.729$ for compaction
and $0.605$ for the memory tools.

The second measurement holds the note fixed and moves only the information in it. For each note we take every
segment whose content words overlap the note's own task context, which is the segment set that carries the task,
and replace that whole set with the corresponding set of a note written for another task under the same
interface, choosing the donor so the two sets have the same character count. Every other byte of the note, and
its call scaffolding, stays as written. The teacher then scores the note as written against the note whose key
information comes from another task, again as the mean log-probability per memory-content token.

Over $60$ notes per interface the teacher ranks the untouched note above the replaced one in \textbf{60 of 60}
cases on the file interface, \textbf{55 of 60} on compaction and \textbf{50 of 60} on the memory tools
(Table~\ref{tab:keyset-swap}), with mean margins of $+0.577$, $+0.528$ and $+0.480$ nats. A one-sided Fisher
test against the file interface gives $p=0.029$ for compaction and $p=0.0007$ for the memory tools.

\begin{table}[t]
\centering
\small
\setlength{\tabcolsep}{3pt}
\begin{tabular}{@{}lcccc@{}}
\toprule
Interface & Correct & 95\% interval & Margin & $p$ vs file \\
\midrule
File & 60/60 & $[0.940,1.000]$ & $+0.577$ & --- \\
Compaction & 55/60 & $[0.816,0.972]$ & $+0.528$ & $0.029$ \\
Memory tools & 50/60 & $[0.715,0.917]$ & $+0.480$ & $0.0007$ \\
\bottomrule
\end{tabular}
\caption{\textbf{Replacing a note's key information by another task's.} Each note is scored against itself
with its key segments replaced, matched in character count. Intervals are exact binomial, the margin is per
memory-content token, and the last column is a one-sided Fisher test against the file interface.}
\label{tab:keyset-swap}
\end{table}

Figure~\ref{fig:readiness-traffic}a measures how surprising each interface is to the two models. The preference
above measures what that surprisal buys: on the interface both models already use, the teacher's score separates
a write that carries the task's information from one that does not, and it separates them less often on the
other two.

\subsection{Memory writes without a required checkpoint}
\label{app:no-required-write}

The policy writes its state at a boundary, and the environment waits for that write before it replaces the
context. Remove the requirement and the replacement proceeds either way, so a policy that wants its state to
survive still has to write before it. Under that setting the trained policy keeps most of its success, both when
it is evaluated as trained and when the same schedule is run in the new environment (Table~\ref{tab:no-force}).
The second row is therefore the stronger statement: the memory behaviour does not need the requirement to be
learned.

\begin{table*}[t]
\centering
\small
\setlength{\tabcolsep}{8pt}
\begin{tabular*}{\textwidth}{@{\extracolsep{\fill}}lcccccc@{}}
\toprule
Method & Shop & Coding & AutoResearch & General & Visual & Avg. \\
\midrule
CAMP-II & 100.0 & 37.5 & 64.1 & 81.2 & 40.2 & 64.6 \\
without the requirement, as trained & 100.0 & 35.9 & 61.7 & 78.1 & 37.5 & 62.6 \\
without the requirement, retrained & 100.0 & 36.7 & 62.5 & 80.5 & 40.2 & 64.0 \\
\bottomrule
\end{tabular*}
\caption{\textbf{Removing the requirement to write at a boundary.} Task success on the CAMG-II test set when
the context is replaced whether or not a valid continuation write exists. The second row evaluates the trained
CAMP-II-4B in that environment, the third trains the same recipe there.}
\label{tab:no-force}
\end{table*}

\subsection{Writes and read-backs across a replacement}
\begin{figure*}[!t]
  \centering
  \includegraphics[width=0.98\textwidth]{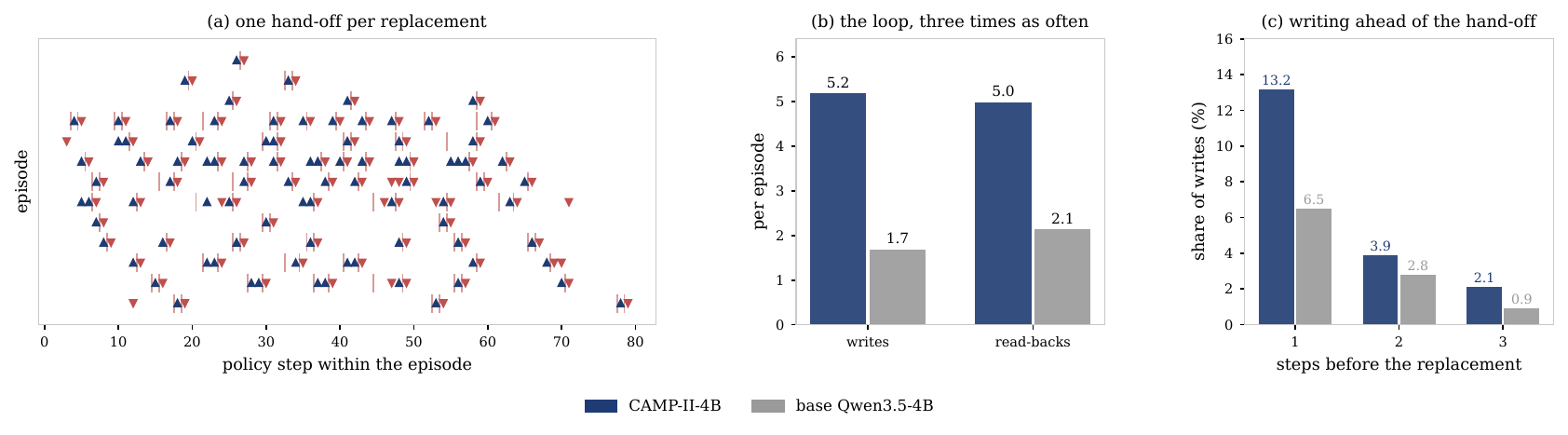}
  \caption{\textbf{A note written before a replacement is read back after it.}
  (a) Writes ($\blacktriangle$), read-backs ($\blacktriangledown$) and context replacements ($\vert$) along
  $14$ CAMP-II-4B Terminal-Bench episodes, one row per episode. (b) Writes and read-backs per episode for
  CAMP-II-4B and for the initial checkpoint on the same episodes. (c) Share of writes that land at least one,
  two or three steps before the replacement.}
  \label{fig:lifecycle}
\end{figure*}

A trained policy writes its note before a replacement and reads it back after, and it does so about three times
as often as the initial checkpoint. We follow the CAMP-II-4B policy through the $65$ Terminal-Bench episodes that ran to completion in our
evaluation run and mark at every step whether it writes the note that carries its continuation, whether it reads it
back, and where the context is replaced. $329$ of the $337$ written notes are read back after exactly one replacement, and $3$ are never read again (Fig.~\ref{fig:lifecycle}a,b). The trained policy writes the file $5.2$ times per episode and reads it back
$5.0$ times, while the initial checkpoint writes it $1.7$ times and reads it back $2.1$. It also starts
earlier: $13.2\%$ of its writes land at least one step before the replacement, against $6.5\%$ for the initial
checkpoint (Fig.~\ref{fig:lifecycle}c).

\section{Extended Related Work}
\label{app:extended-related-work}

This appendix covers families beyond the three surveyed in the main text. The memory-aware OPD
methods compared experimentally are introduced in Sec.~\ref{sec:setup}.

\paragraph{Context Compression.}
What enters the active context can itself be learned, at several different sites.
\citet{wu2025resum} condense history with an external summarizer and train the policy to reason
over the summaries, propagating reward across segmented trajectories. A separate compressor can
be distilled from optimized compression guidelines~\citep{kang2026acon}, and a memory-view
router can be trained by supervised distillation followed by grouped GRPO while the acting
policy stays frozen~\citep{chen2026memprism}. Cartridges moves the compression into the
key--value cache: corpus-specific caches are trained offline from synthetic conversations and
context distillation with the language model frozen~\citep{eyuboglu2026cartridges}. CAMP-II
keeps memory as ordinary files and bounds only the continuation file, so compression is one
file operation among others rather than the memory representation itself.

\paragraph{Long-Horizon Agent Environments.}
These environments provide task feedback over extended agent interactions: existing
benchmarks span heterogeneous interactive tasks~\citep{liu2024agentbench}, and training
frameworks support multi-turn policy optimization~\citep{xi2026agentgymrl}. CAMG-II contributes
five heterogeneous long-horizon environments, including visually grounded web development, for
joint memory post-training through a shared file-memory interface~\citep{luo2026camg}.

\paragraph{Experience Reuse.}
Experience reuse conditions learning on what earlier episodes produced.
\citet{zhang2026opdevolver} distill the selection, use, writing, and maintenance of a
hierarchical experience repository, and \citet{ye2026opcd,ye2026oel} condition distillation on
historical solution traces and guiding contexts, with the latter iterating experience
extraction and consolidation. A second group builds self-teachers from experience:
cross-problem behavioral patterns serve as a memory-conditioned scaffold that is removed at
inference~\citep{liu2026pmd}, skill-conditioned self-distillation is gated alongside task
RL~\citep{lu2026sdar}, and hindsight skills are extracted as the supervising
signal~\citep{yang2026opid}. \citet{wu2026seed} couple policy updates with an evolving
trajectory analyzer, \citet{tu2026ucob} select the local teaching direction by comparing
skill-conditioned and plain-context returns, and BCSD augments and prunes explicit skill
context to rescale RL advantages, retaining skill guidance at inference~\citep{pan2026bcsd}.
These memories support reuse and hindsight-guided learning across tasks, whereas CAMP-II
retains only the current episode's state for continued execution.

\paragraph{On-Policy Distillation.}
The design space of OPD~\citep{agarwal2024opd} extends beyond the memory setting. The teacher is
one design choice: \citet{zhao2026self} construct a self-teacher from privileged information
and fix the initial policy as the teacher, while \citet{hubotter2026sdpo} condition a
self-teacher on rich textual feedback or successful rollouts and use an exponential-moving-
average one. \citet{li2026rethinking} analyze the mechanisms and practical recipe of OPD.
Filtered privileged rollouts can be combined with RLVR to address prompts yielding zero
reward~\citep{ding2026hdpo}, and complete student trajectories can adapt the teacher while
prefix-only student inputs are preserved~\citep{feng2026past}. The trajectory can also be
repaired before the supervision is applied: unreliable token supervision can be attributed to
failed prefixes and the rollout revised~\citep{jiang2026trajectory}, teacher--student
continuation asymmetry can trigger a short teacher leg~\citep{relayopd2026}, and an
intervention can be validated through its effect on the student's
continuation~\citep{futurebridgeopd2026}. The training objective, teacher parameters, and
scoring information are separate design choices. CAMP-II holds the teacher fixed and varies
the training objective within the same file-memory interface.

\paragraph{Token-Level Supervision.}
A separate axis within OPD selects which of the tokens already produced should carry
supervision. Privileged action traces can change the teacher's input while the student
objective still spans responses~\citep{penaloza2026privileged}, and restricting privileged
input is a different choice from restricting the distillation loss to action tokens.
Supervision can vary by within-sequence position~\citep{liu2026when}. A teacher-directed
perturbation's first-order effect on the RL surrogate can redistribute auxiliary-loss mass
within a turn~\citep{icsd2026}. Token selection and privileged-information adjustment can be
coupled to a student learning-difficulty budget~\citep{yang2026capacity}. A reference-only
component can be subtracted to construct a question-conditioned
target~\citep{shen2026purified}. Supervision can be added on aligned final-layer hidden
states, projecting the teacher--student residual through a frozen rank-64 factor estimated
from residual covariance and LM-head Fisher sensitivity~\citep{scopeopsd2026}.
Tutor consensus can admit supervision when no external correctness labels are
available~\citep{stein2026gates}. These approaches select supervision by reliability,
estimated utility, or representation geometry. A next-token score can be myopic about the
outcome even for a capable teacher~\citep{myopic-scorer,when-to-follow}, and the overlap
between the two models predicts whether on-policy sampling helps at all~\citep{overlap-select}.
Which tokens a memory task produces at all is decided upstream, by the interface. CAMP-II
fixes that upstream choice and leaves the per-token criterion to these methods.

\paragraph{Distillation with Privileged Information.}
Teacher supervision often rests on information the student will not have at deployment.
Policy Distillation transfers control policies into a student~\citep{rusu2016policy}, and
generalized distillation lets the supervision be computed from training-only privileged
information~\citep{lopezpaz2016unifying}. In partially observable control, privileged state
supports provably efficient learning under specified structural
assumptions~\citep{cai2024provable}, but asymmetry also creates limits: a teacher's
state-conditioned decision may be unrealizable from the student's
observations~\citep{kim2025realizable}, the relative merits of distillation and direct
decision learning depend on the problem class~\citep{song2025distill}, and
\citet{jiang2026asymmetric} train a smaller critic to supervise a fixed proprietary actor, so
oversight comes from a model that sees less than the actor rather than one that sees more.
Whether privileged content actually helps admits causal controls. Replacing a paired solution
with another problem's solution retains substantial gains, exposing context-induced teacher
behavior~\citep{ichihara2026privileged}. Correct-reference benefits vary with generation
mode, scale, and dataset~\citep{shrestha2026rethinking}. Reference value varies across rollout
configurations when multiple reasoning views are compared with reference-free
distillation~\citep{zhang2026what}. Changes in token likelihood are distinct from changes in
expected task outcome~\citep{nguyen2026privileged}.
Self-distillation as a lone objective
can show falling loss without corresponding validation gains~\citep{harne2026privileged}.
These findings separate target fit, reference utility, and deployable task behavior, and
CAMP-II measures those links with content-target probes, memory interventions, and test-set
evaluation. The frozen teacher scores only what the student can see, and only the student
remains afterwards.

\paragraph{Agent Distillation.}
Agent distillation transfers a teacher's task-solving capability into a smaller deployable
student. \citet{kang2025distilling} transfer full task-solving behavior with retrieval and
code tools through teacher trajectories, and \citet{lyu2026student} extend the recipe with
student rollouts: behavior cloning of the trajectories first, then teacher correction and
further fine-tuning, then short-horizon RL. At the inference-time end, Agent Memory
Distillation transfers hierarchical teacher memories through prompting and retrieval without
parameter training~\citep{kim2026amd}. Compact skills summarized from agent trajectories can
serve as dynamic privileged information~\citep{wang2026skill}, and RetireOPD trains a
decoupled skill-conditioned teacher whose supervision is retired once student--teacher
discrepancy and success criteria are met~\citep{yu2026retireopd}. What transfers is the
teacher's task behavior, and when its supervision stops is a design choice. The complementary
question in CAMP-II is how the memory interface affects supervision of the acting policy's
own task-state records.

\paragraph{Behavioral Side Effects of Distillation.}
Distillation can change behavior beyond task success. Sampled-demonstration self-distillation
can concentrate correct outputs and flatten pass@$k$ gains~\citep{nicolicioiu2026diversity},
and \citet{peng2026thinking} diagnose thinking collapse through suppression of epistemic
tokens at high-entropy decision points. Self-distillation has also been studied as a mechanism
for continual learning~\citep{shenfeld2026continual}, whereas Denser $\neq$ Better finds
stronger forgetting and formatting feedback loops in its tested continual
setting~\citep{wang2026denser}. When Context Returns studies a different failure: performance
can deteriorate when the privileged context is restored, and no-context anchoring can reduce
this harm~\citep{wang2026contextreturns}. Together these studies motivate measuring reasoning
behavior, diversity, and context sensitivity alongside success. CAMP-II evaluates task success
together with paired content interventions, and leaves these behavioral measures outside its
evaluation.

\end{document}